\documentclass{article}

\PassOptionsToPackage{numbers}{natbib}

\usepackage[preprint]{neurips_2026}

\usepackage[utf8]{inputenc} 
\usepackage[T1]{fontenc}    
\usepackage{hyperref}       
\usepackage{url}            
\usepackage{booktabs}       
\usepackage{amsfonts}       
\usepackage{nicefrac}       
\usepackage{microtype}      
\usepackage{xcolor}         

\usepackage{algorithm2e}
\usepackage{amsmath}
\usepackage{graphicx}
\usepackage{multicol}
\usepackage{multirow}
\usepackage{tikz}
\usetikzlibrary{positioning, calc, shapes.geometric, arrows.meta, backgrounds, fit, decorations.pathreplacing, 3d}
\usepackage{wrapfig}

\title{NAPS: Attention-Based Fusion of Heterogeneous Physiological Signals}

\author{Alvise Dei Rossi \\
  Faculty of Informatics, Università della Svizzera Italiana \\
  Inst. of Digital Tech. for Personalized Healthcare, SUPSI \\
  Lugano, Switzerland \\
  \texttt{alvise.dei.rossi@usi.ch} \\
  \AND
  Julia van der Meer \& Markus H. Schmidt \& Claudio L.A. Bassetti \\
  Sleep Wake Epilepsy Center, Department of Neurology, Inselspital \\
  University of Bern \\
  Bern, Switzerland \\
  \texttt{\{julia.vandermeer, markus.schmidt, claudio.bassetti\}@insel.ch} \\
  \And
  Luigi Fiorillo \\
  Inst. of Digital Tech. for Personalized Healthcare, SUPSI \\
  Ente Ospedaliero Cantonale (EOC) \\
  Lugano, Switzerland \\
  \texttt{luigi.fiorillo@supsi.ch} \\
  \And
  Silvia Santini \\
  Faculty of Informatics \\
  Università della Svizzera Italiana \\
  Lugano, Switzerland \\
  \texttt{silvia.santini@usi.ch} \\
  \And
  Francesca Faraci \\
  Inst. of Digital Tech. for Personalized Healthcare, SUPSI \\
  Lugano, Switzerland \\
  \texttt{francesca.faraci@supsi.ch}
}

\begin{document}

\maketitle

\begin{abstract}
    Physiological signals are inherently heterogeneous: they are collected under diverse acquisition setups, differ in the number and type of modalities and channels, varying in quality, reliability, and relevance across tasks. 
    This variability poses a major challenge for machine learning models required to generalize across subjects, sensors, and clinical environments. Existing approaches typically train on limited modalities or single channels, leading to marginal representations that, on their own, fail to capture the systemic complexity of the physiological state; na\"ive fusion of such representations, such as via pooling or voting schemes, is typically suboptimal, as it cannot adaptively weight different sources or capture temporal, spatial, and cross-modality dependencies.
    We introduce \textbf{NAPS} (\textbf{N}eural \textbf{A}ggregator of \textbf{P}hysiological \textbf{S}ignals), a neural module that performs principled data fusion to derive unified physiological representations, employing an ad hoc tri-axial attention mechanism and dimension-adaptive training to robustly manage varying high-dimensional sensor configurations.
    We test NAPS on automatic sleep staging from polysomnography (PSG), an ideal real-world application, where recordings consist of multiple physiological signals (EEG, EOG, EMG, $\ldots$), considerably varying in configuration across datasets and institutions. 
    Leveraging frozen pretrained unimodal encoders, NAPS dynamically integrates representations or predictions, achieving state-of-the-art generalization across multiple datasets. 
\end{abstract}

\section{Introduction}
\label{sec:intro}

The integration of multiple data types, spanning text, images, and signals, has emerged as a rapidly growing frontier in machine learning \cite{Alayrac2022, Jin2024, Wu2024}.
This paradigm is particularly critical in modern healthcare, which is shifting from episodic diagnostics to continuous, pervasive monitoring \cite{Matwyshyn2019}, and where clinical decision-making relies on evidence drawn from diverse sources such as medical imaging, electronic health records, and physiological time series \cite{Acosta2022}.
While these sources provide complementary perspectives on an individual’s health status, they also introduce substantial heterogeneity, differing in format, acquisition protocols, and clinical relevance across tasks \cite{Krones2025, Thapa2025}. Importantly, this heterogeneity arises not only between data types but also within data types. 
Combining information from (sub-)networks that process distinct modalities \cite{Kline2022}, either by fusing representations or predictions \cite{Stahlschmidt2022}, holds considerable promise for improving downstream performance. 
Models trained on a subset of inputs (e.g., a single modality or channel) are intuitively suboptimal when richer multimodal data are available \cite{Phan2022Review}.
Managing this complexity is essential for developing generalizable models, yet presents significant methodological difficulties \cite{Wang2020_multi, Huang2022}.

In this work, we investigate data fusion in the context of multivariate time series, specifically focusing on the challenge of obtaining unified representations from multimodal signals acquired from varying sensor layouts. We focus on the scenario where pre-trained unimodal models are available but must be integrated dynamically. We evaluate our approach on overnight polysomnography (PSG) recordings, the clinical gold standard for diagnosing sleep–wake disorders \cite{Christensen2015, Ibanez2018}, which provide an appropriate testbed for multi-sensor integration.
PSG exemplifies the intrinsic variability of physiological signals: recordings combine several modalities (e.g., EEG, EOG, EMG, $\ldots$) collected using channel configurations that differ substantially across clinical centers \cite{Phan2022Review}. These signals are typically segmented into 30-second windows, referred to as \emph{sleep epochs}, and manually classified into five sleep stages (Wake, N1, N2, N3, REM) according to standardized scoring criteria \cite{Berry2017}.

Manual sleep staging is labor-intensive and time-consuming, limiting its scalability for addressing the diagnostic needs of the millions affected by sleep-wake disorders worldwide \cite{Altevogt2006}. 
As a consequence, a growing body of research has explored machine learning–based approaches to automate sleep staging, aiming to improve its efficiency and accessibility \cite{Phan2022Review}.
Advances in deep learning, coupled with the growing availability of large, annotated PSG datasets \cite{Zhang2024} and with the emphasis on large-scale multi-cohort training \cite{Olesen2021}, have allowed recently proposed models to achieve robust zero-shot performance across heterogeneous clinical and research settings \cite{Perslev2021, DeiRossi2025}. 
We note that several advancements toward this goal so far have been achieved making use of straightforward ensembling techniques to handle data heterogeneity, such as (soft-)voting across channels \cite{Perslev2021} and/or models \cite{Stephansen2018, DeiRossi2025}. 
While such approaches are appealing due to their simplicity and modularity, readily accommodating varying numbers of modalities, channels, and architectures, making them broadly applicable across diverse input configurations, we argue that they entail significant limitations.
First, voting mechanisms assume that averaging is an adequate function for combining representations or predictions across channels, modalities, and possibly different modeling paradigms. 
Second, these approaches assign fixed importance, usually equal, to all contributors. This rigidity is detrimental when several channels suffer from poor signal quality or when a modality with inherently low predictive power is included. 
Finally, in sequence-to-sequence settings, voting-based fusion typically operates at segment level, disregarding temporal dependencies that could otherwise be exploited.
To address such limitations while retaining modularity, we make the following contributions:

\begin{itemize}
    \item We propose \textbf{NAPS} (\textbf{N}eural \textbf{A}ggregator of \textbf{P}hysiological \textbf{S}ignals), an ad hoc attention-based meta-model for multivariate time series, which learns to aggregate representations from multiple pretrained single-channel encoders by capturing temporal, spatial, cross-view, and cross-modality dependencies.
    \item We generalize criss-cross attention \cite{Huang2019} beyond spatio-temporal dimensions \cite{Wang2024} to a tri-axial attention mechanism, explicitly modeling segment-level dependencies across multiple representations of the original signal.
    \item We propose a dynamic batch sampling protocol to handle varying modalities, number of channels, encoders, and sequence lengths, both at training and inference time, avoiding any masking or padding and promoting generalization across diverse input data configurations.
    \item We test NAPS in both intermediate (feature-level) and late (prediction-level) fusion configurations for automatic sleep staging, achieving state-of-the-art zero-shot performance on multiple datasets, both in-domain and out-of-domain. NAPS is engineered for out-of-the-box deployment. For end-users, it operates entirely zero-shot, requiring no site-specific fine-tuning.
\end{itemize}

\section{Related work}

We ground our work in the multimodal deep learning taxonomy proposed by \cite{Stahlschmidt2022}. In this framework, \emph{joint representations} capture latent factors shared across modalities, whereas \emph{marginal representations} refer to the transformed outputs derived from unimodal input data. We expand this notion by referring to marginal representations as conditional representations derived from any subset of the original multimodal input data.
In \emph{early fusion}, raw multimodal inputs are concatenated and processed jointly as a single entity, producing joint representations directly. 
This approach enables the simultaneous learning of within-modality and cross-modality dependencies but does not disentangle modality-specific factors. By contrast, in \emph{intermediate fusion}, sub-networks first extract marginal representations, which are then fused into joint representations (e.g., via attention-based mechanisms \cite{Chen2020}). 
Finally, in \emph{late fusion}, integration occurs at the decision level; this facilitates combining deep and shallow methods and offers high flexibility, but omits the explicit modeling of latent interactions. 
Approaches building upon marginal representations can offer strong modularity advantages, as pre-trained modality-specific networks, potentially trained on entirely different datasets or only partially overlapping ones, can serve as feature extractors for marginal embeddings, which can subsequently be integrated by flexible fusion modules. Moreover, leveraging marginal representations from pre-trained networks, further supported by training strategies promoting generalization across modalities, can circumvent the challenge of modality competition \cite{Wang2020_multi, Huang2022}. 

Integration of information across modalities and channels is particularly relevant in the context of PSG data modeling \cite{Phan2022Review}. While a large number of studies have focused on models leveraging electroencephalography (EEG), electrooculography (EOG), and chin electromyography (EMG) \cite{Fiorillo2019}, typically included in PSG recordings, other modalities have been explored, including electrocardiography (ECG) \cite{Jones2024}, photoplethysmography \cite{Radha2021}, actigraphy \cite{Zhai2020}, respiratory signals \cite{Bakker2021}, audio \cite{Dafna2018}, video \cite{Cesari2022}, and unusual derivations such as sternocleidomastoid EMG~\cite{VanGorp2025}. Although EEG-only models can match inter-rater agreement in sleep staging, complementary modalities provide a holistic view of sleep dynamics~\cite{Phan2022Review} and may offer insights into a patient's health status beyond sleep staging~\cite{Thapa2025}.

From a methodological standpoint, traditional feature-engineering approaches \cite{Vallat2021} have largely been superseded by deep learning methods enabling end-to-end representation learning \cite{Phan2022Review}. 
Given the systematic trends in human sleep cycles \cite{Feinberg1979} and the American Academy of Sleep Medicine (AASM) guidelines recommending clinicians to consider surrounding epochs when scoring sleep stages \cite{Berry2017}, modern architectures employ sequence-to-sequence modeling with bidirectional context to capture the temporal dependencies essential for accurate classification \cite{Phan2019}. 
Proposed approaches span diverse input representations, from raw time-series signals to time–frequency representations, and modeling paradigms including convolutional \cite{Perslev2021}, recurrent \cite{Phan2019, Phan2023}, hybrid \cite{Olesen2021}, and attention-based \cite{Phan2022sleeptransformer} networks. 
Several efficient transformer variants have been developed to address the quadratic complexity of self-attention, enabling scalable modeling of long biomedical sequences \cite{Fournier2023, Wang2020, Yang2023, Jaegle2021}. Notably, a recent adaptation of criss-cross attention \cite{Huang2019} has outperformed state-of-the-art methods on several EEG tasks by explicitly modeling spatio-temporal dependencies via disentangled attention pathways \cite{Wang2024}, demonstrating that such inductive bias is valuable in physiological signal modeling.

Existing approaches vary significantly in how they handle sensor heterogeneity. Early studies predominantly adopted fixed early fusion configurations, restricted to a predefined subset of modalities and specific channel derivations. Such rigidity, however, poses major challenges when encountering missing channels or variations in modalities at test time \cite{Rayan2024}. Recent studies have increasingly acknowledged these limitations. EEG foundation models such as LaBraM \cite{Jiang2024}, CBraMod \cite{Wang2024}, and Luna \cite{Doner2025} have introduced architectures adaptable to arbitrary numbers of electrodes, although their primary applications remain within the brain computer interface domain. Flexibility at modality-level, considering variable availability of EEG and EOG signals, but fixed channel configurations, was recently explored in \cite{Kontras2024}, where end-to-end unimodal encoders were jointly trained and combined through cross-attention and multi-loss mechanisms to yield a multimodal system which outperformed its unimodal counterparts.
Other large-scale multimodal frameworks such as 
SleepFM \cite{Thapa2025} have shown promising flexibility in integrating variable input structures across both modalities and channels, through clever combinations of masking, padding, pooling, and embedding mechanisms.
SleepFM implements an intermediate fusion design, in which homogeneous 1D convolutional encoders process each $<$modality, channel$>$ pair separately, followed by transformer blocks that contextualize representations prior to unweighted averaging across spatial, temporal, and finally modality dimensions, yielding marginal representations at different levels of the hierarchy \cite{Thapa2025}. 
Another influential line of work originated with U-Sleep \cite{Perslev2021}, which leveraged a majority-voting scheme across all channel combinations involving any single EEG and EOG derivation. In practice, this constitutes an hybrid fusion approach: while modalities integration occurs early, integration across channel combinations takes place at the decision stage. Remarkably, this approach was agnostic to the specific channel derivations, their placement, and referencing schemes \cite{Fiorillo2023}. Such flexibility was enabled by a training strategy that generated batches through random sampling of the channels based on their availability.
A similar idea of enforcing channel flexibility during training was concurrently proposed in the DAT framework \cite{Malekzadeh2021}. Variants of it, such as random lead selection, have also been recently proposed \cite{Monachino2025}.
Building upon these ideas, SLEEPYLAND \cite{DeiRossi2025}, adopted the training strategy of \cite{Perslev2021} on a larger scale, uniformly training diverse high-performing architectures, namely U-Sleep \cite{Perslev2021}, DeepResNet \cite{Olesen2021}, and SleepTransformer \cite{Phan2022sleeptransformer}, configured for single-channel EEG, EOG, and their combinations. 
Comprehensive benchmarking revealed that no single model was clearly superior; instead simple soft-voting across channels, modalities, and models offered the most reliable out-of-domain performance across varied evaluation conditions.

\begin{figure*}[!t]
    \centering
    \includegraphics[width=.84\linewidth]{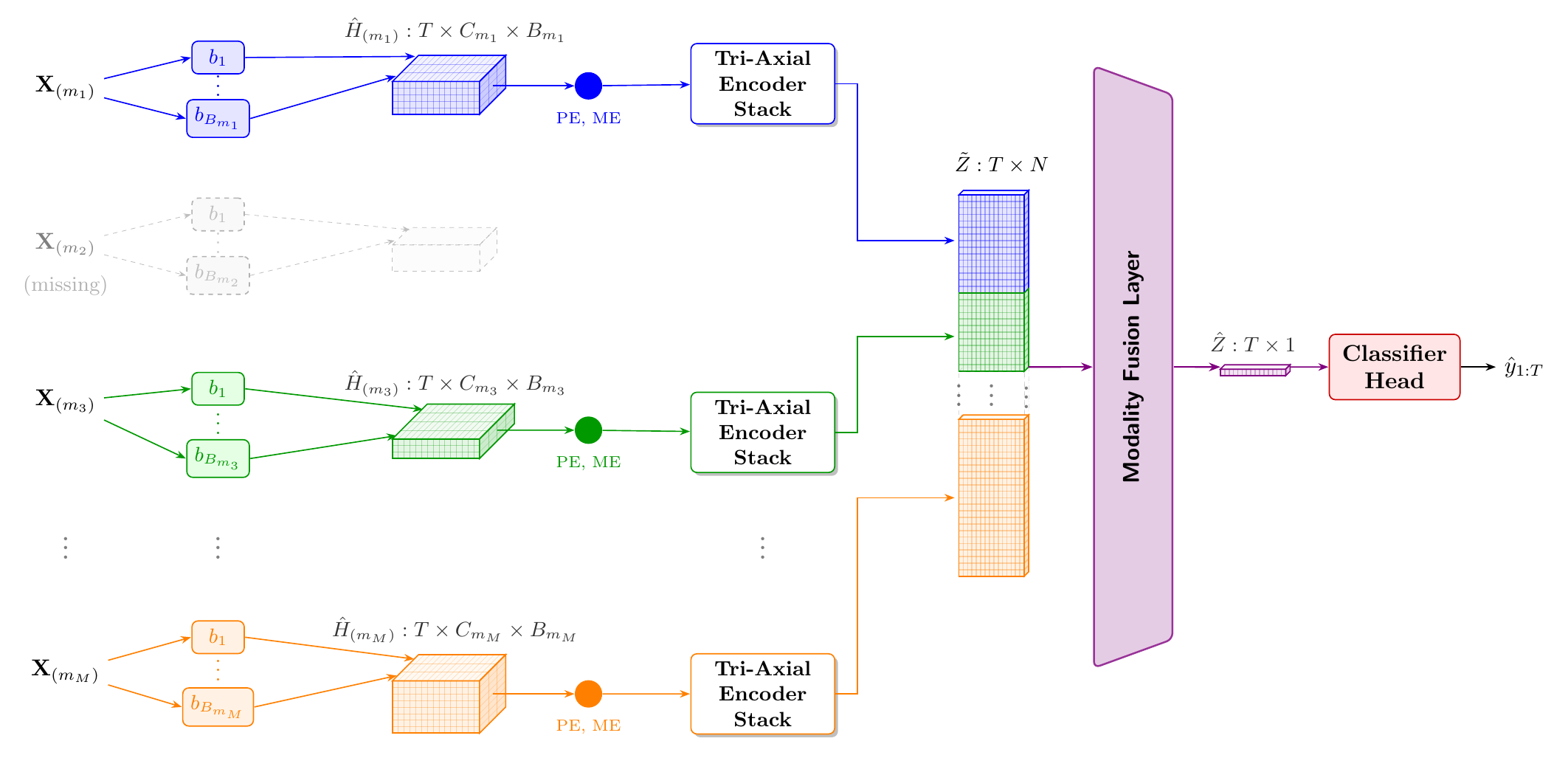}
    \caption{Overview of the NAPS architecture. The geometry of the latent tensors $\hat{H}_{m_k}$ reflects the input structure, where width, height, and depth correspond to Time ($T$), Channels ($C_{m_k}$), and Base Encoders ($B_{m_k}$) (batch and feature dimensions are omitted for clarity). PE: positional encoding, ME: modality encoding. The module is adaptable to diverse input dimensions and to missing modalities.
    }
    \label{NAP_architecture}
\end{figure*}

\section{Methods}

\subsection{Model architecture}
The NAPS architecture, shown in Figure \ref{NAP_architecture}, implements a principled fusion approach for temporally-aligned multivariate time series. Specifically, we address the case in which the multivariate structure of a time series arises from the concurrent recording of multiple modalities, each comprising one or more spatial channels—that is, distinct sensor measurements providing complementary, partially overlapping, perspectives on the same underlying process. Moreover, for each $<$modality, channel$>$ pair, multiple representations may be available, as each signal can be processed by models emphasizing alternative \emph{views}, capturing specific aspects of the original input data \cite{Cohen1995}, such as the time domain, time–frequency domain, or handcrafted features, or be based on diverse modeling paradigms from the same domain. We call this third dimension \emph{blending dimension}, to emphasize that we're blending multi-view representations. We note that using representations from multiple views can be crucial, since the informational value of different feature domains is often task-specific. Allowing the model to adaptively balance which view is most relevant for a given task is thus beneficial \cite{Phan2021, Ahamed2025}.

NAPS leverages a set of pre-trained, unimodal, single-channel frozen base models, to obtain marginal representations and can be used for either intermediate or late fusion, depending on the outputs of the base models. 
NAPS features four distinct sub-modules: (i) a \emph{base generator}, producing per-$<$modality, channel$>$ marginal representations or predictions from the available base encoders; (ii) a \emph{tri-axial attention encoder}, which mixes information along temporal, spatial, and blending dimensions; (iii) a \emph{modality fusion layer}, responsible for attention-based integration of information across modalities; and (iv) a \emph{classifier head} that yields the probability distributions. 
In the following paragraphs we detail each module in a sequence-to-sequence multi-class classification scenario, assuming all available input sources are used and focusing on a single instance within a batch.

\paragraph{Base generator}
Let a multivariate time series $X$ be represented as a sequence of $T$ contiguous segments of the same length, $(\mathbf{x}_1, \ldots, \mathbf{x}_T)$, each associated with a ground-truth label $y_t \in \mathcal{S}$. 
We denote by $M$ the number of modalities in $X$, where each modality $m_k$ has $C_{m_k}$ available channels and $B_{m_k}$ associated pre-trained encoders.
For a modality $m_k$ ($k=1,\ldots,M$), channel $c_j$ ($j=1,\ldots,C_{m_k}$), and base encoder $b_\ell$ ($\ell=1,\ldots,B_{m_k}$), the corresponding output is:

\begin{equation}
    \mathbf{\hat{h}}_{(m_k, c_j, b_\ell)}=\left\{ \mathbf{\hat{h}}_{(m_k, c_j, b_\ell),t} \;\big|\; t=1,\ldots,T \right\}
\end{equation}
where $\mathbf{\hat{h}}_{(m_k, c_j, b_\ell),t}$ is either a segment-level representation or a task-specific output vector $\in \mathbb{R}^{|\mathcal{S}|}$, depending on whether intermediate or late fusion is applied. 
For instance, for PSG and automatic sleep staging, segments are sleep epochs of 30 seconds, $ \mathcal{S} = \{\mathrm{Wake}, \mathrm{N1}, \mathrm{N2}, \mathrm{N3}, \mathrm{REM}\}$, and $\mathbf{\hat{h}}_{(m_k, c_j, b_\ell),t}$ represents a sleep epoch embedding or a probability distribution over $\mathcal{S}$. 
The set of all $\mathbf{\hat{h}}_{(m_k, c_j, b_\ell),t}$ is then linearly projected into a feature space $\in \mathbb{R}^{d_{\text{model}}}$.

\begin{wrapfigure}{r}{0.4\textwidth} 
    \centering
    \includegraphics[width=\linewidth]{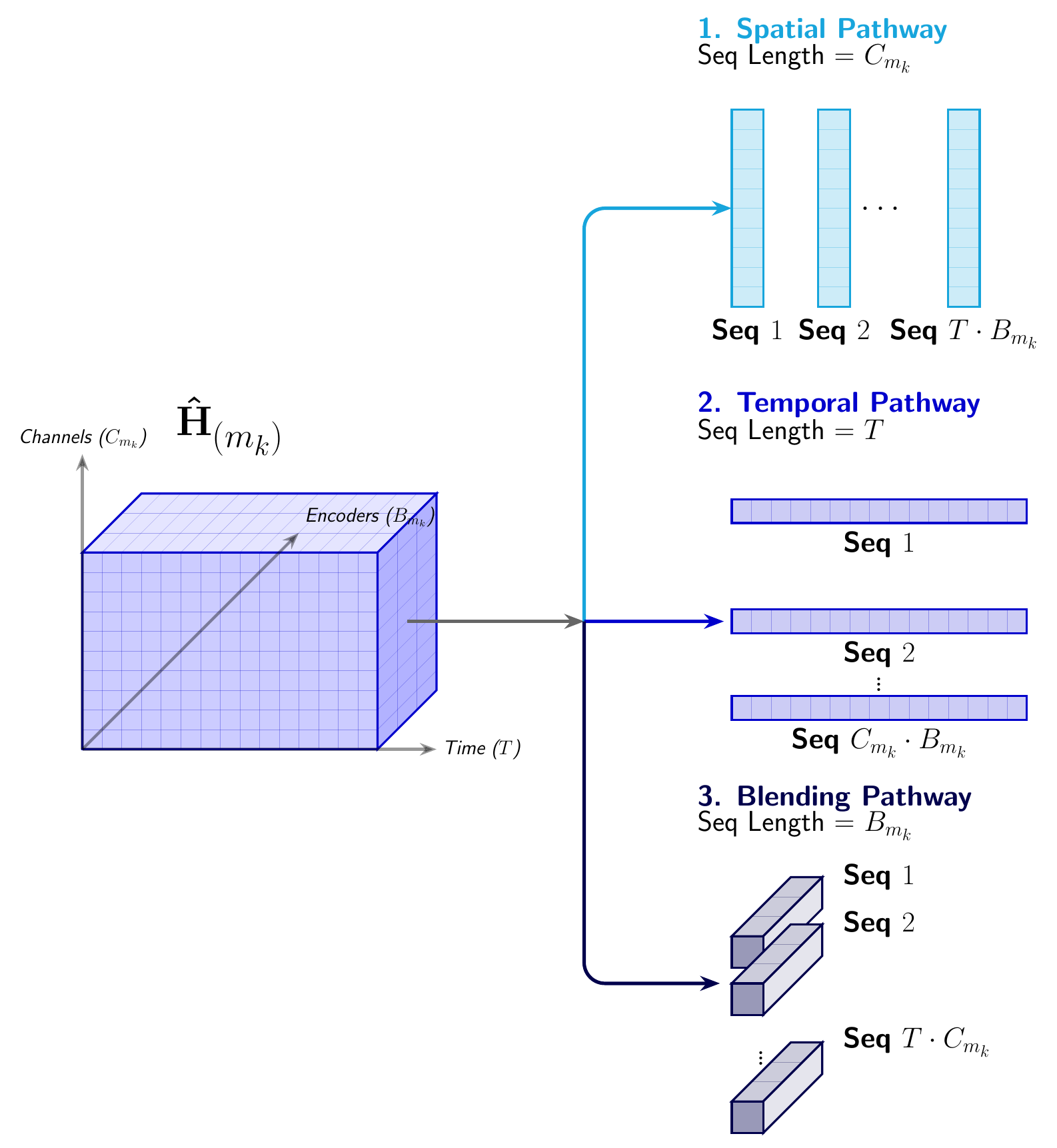}
    \caption{In tri-axial attention, three pathways process sequences along distinct axes, utilizing $h/3$ dedicated heads for each.}
    \label{fig:tri_axial_reshaping}
\end{wrapfigure}

\paragraph{Tri-axial attention encoder}
For a single modality $m_k$, the projected features are collected in a tensor $\mathbf{\hat{H}}_{(m_k)} \in \mathbb{R}^{T \times C_{m_k} \times B_{m_k} \times d_{model}}$.
To encode temporal order, we add the standard sinusoidal positional encoding (PE) from \cite{Vaswani2017}. 
Furthermore, to provide modality identity, we employ a learnable modality embedding (ME) vector uniquely assigned to $m_k$ \cite{Jaegle2021}.
The resulting tensor is then processed by $L$ stacked transformer encoder layers, employing a tri-axial self-attention mechanism that extends the criss-cross attention paradigm \cite{Huang2019, Wang2024}. Instead of computing a single joint attention map over all dimensions, the mechanism decomposes multi-head attention into three pathways, each attending along a different axis of the input tensor:

\begin{itemize}
    \item \textbf{Spatial attention:} Attends along the spatial axis while holding the temporal and blending dimensions fixed, capturing cross-channel dependencies for all $T\cdot B_{m_k}$ spatial sequences of length $C_{m_k}$ in parallel. 
    \item \textbf{Temporal attention:} Attends along the temporal axis while keeping spatial and blending dimensions fixed, enabling NAPS to learn temporal dependencies for all $C_{m_k} \cdot B_{m_k}$ temporal sequences of length $T$ in parallel.
    \item \textbf{Blending attention:} Attends along the blending axis while keeping temporal and spatial dimensions fixed, facilitating the fusion of segment representations for all $T \cdot C_{m_k}$ blending sequences of length $B_{m_k}$ in parallel.
\end{itemize}

A sketch of the proposed structure is proposed in Figure \ref{fig:tri_axial_reshaping}. The $h$ attention heads are divided evenly across the three pathways, allowing each group of $h/3$ heads to specialize in modeling dependencies along a single axis. 
In attention computations we apply query/key normalization \cite{Ba2016} before the scaled dot-product, which stabilizes training and improves convergence, and omit the bias term to speed up training \cite{Dehghani2023, Jiang2024}. For the spatial pathway, the attention output is computed as:

\begin{equation}
    Z_s^{(i)} = \mathrm{Softmax}\!\left(\frac{\mathrm{LN}(Q_s^{(i)}) \, \mathrm{LN}(K_s^{(i)})^\top}{\sqrt{d_{\text{k}}}}\right) V_s^{(i)}
\end{equation}

\begin{equation}
    Z_s = \mathrm{Concat}\left(Z_s^{(1)}, \ldots, Z_s^{(h/3)}\right)
\end{equation}

Analogous multi-head computations produce the temporal and blending pathway outputs $Z_T$ and $Z_B$. While the three pathways are logically separated, the initial projection is shared, computation for the different heads within the same pathway are carried out in parallel, while attention computations for different pathways are sequential. The three pathways outputs are then concatenated along the feature dimension and passed through a linear projection. 

The tri-axial design offers a substantial computational advantage over full self-attention. Whereas full self-attention over the flattened tensor would incur a quadratic cost of $\mathcal{O}((T C_{m_k} B_{m_k})^{2})$, the tri-axial complexity is $\mathcal{O}(T B_{m_k} C_{m_k}^{2} + C_{m_k} B_{m_k} T^{2} + T C_{m_k} B_{m_k}^{2})$, i.e., quadratic only along a single axis at a time.
This efficiency is critical in high-density sensing or when considering long contexts (more details and considerations on overhead computations are reported in Appendix \ref{apd:A_complexity}).
Finally, a pointwise feedforward network with residual connections and dropout is applied \cite{Vaswani2017}. Throughout the encoder blocks, layer normalization is applied inside the residual connection, as proposed in \cite{Xiong2020}.

\paragraph{Modality fusion layer}
Following the independent processing of each modality $m_k$, the output tensors are concatenated
yielding $\mathbf{\tilde{Z}} \in \mathbb{R}^{T \times N \times d_{\text{model}}}$,
where $N = \sum_{k=1}^M (C_{m_k} \cdot B_{m_k}) $ is the cumulative count of $<$channel, view$>$ pairs across modalities.
To reduce $\mathbf{\tilde{Z}}$ to a compact embedding, we employ an attention-based fusion mechanism \cite{Phan2022sleeptransformer, Bahdanau2014} that learns to weight the contributions of different sensor streams. For each time step $t$, the fusion layer computes a convex combination $\mathbf{\hat{z}}_t = \sum_{n=1}^{N} \alpha_{t,n} \, \mathbf{\tilde{z}}_{t,n}$
where $\alpha_{t,n} \in [0,1]$ are attention weights, obtained by projecting each $\mathbf{\tilde{z}}_{t,n}$ into a space of dimension $d_{\text{A}}$ using a learned transformation, then scored by a trainable 
context vector: 

\begin{equation}
    \alpha_{t,n} = \frac{\exp(\tanh\!\left(W_A \mathbf{\tilde{z}}_{t,n} + b_A\right)^\top u_A)}{\sum_{j=1}^{N} \exp(\tanh\!\left(W_A \mathbf{\tilde{z}}_{t,j} + b_A\right)^\top u_A)}
\end{equation}

where $W_A \in \mathbb{R}^{d_{\text{A}} \times d_{\text{model}}}$, $b_A \in \mathbb{R}^{d_{\text{A}}}$, and $u_A \in \mathbb{R}^{d_{\text{A}}}$. 
Beyond the flexibility of learnable representation weighting, this formulation inherently yields a transparent measure of the model's reliance on specific channels and modalities, enabling explicit, epoch-by-epoch interpretability.

\paragraph{Classifier head}
The segment-level representations are finally fed into a compact classifier head, comprised of a single hidden layer feedforward network, that maps them into task-specific outputs $\hat{y}_{1:T}$. NAPS is trained end-to-end using the cross-entropy loss against the ground-truth labels.

\subsection{Training protocol}

We train NAPS on inputs of varying dimensionality, pushing it to operate across different modality subsets, channel counts, and sequence lengths. This is done by leveraging: 

\paragraph{Dynamic batch sampling}
Batches are generated by randomly selecting a consistent subset of dimensions along four axes: the number of time steps, the set of modalities, the channels within each selected modality, and the set of base encoders. Along the temporal axis, we sample uniformly $K$ sequences of the same random length from each of the $B$ recordings within the batch. A subset of available modalities is then randomly selected and, within each chosen modality, a random subset of channels is independently sampled, allowing the number of selected channels to differ across modalities. Finally, a random subset of base encoders is sampled. Further details are reported in Appendix \ref{apd:A_batching}.
This procedure yields samples which share the same dimensionality within a batch while it may vary between batches. As a result, padding and masking are not required, favoring computational efficiency and low memory overhead. 

\paragraph{Gradient accumulation}
We accumulate gradients over $G$ distinct batches; each optimizer step considers $G \cdot B \cdot K$ sequences.
By combining dynamic batch construction with gradient accumulation, the model is systematically exposed to a diverse set of input configurations within each optimization step, enhancing robustness to heterogeneous and variable-dimensional data \cite{Malekzadeh2021}.

We report in \ref{apd:A_hparams} implementation details of the architecture and training protocol for our use case. 

\subsection{Experiments}
We leverage the open-weights pre-trained single-channel models of SLEEPYLAND \cite{DeiRossi2025} as base encoders, yielding marginal representations for each $<$modality, channel$>$ pair. These models consider the EEG and EOG modalities and include U-Sleep \cite{Perslev2021}, DeepResNet \cite{Olesen2021}, and SleepTransformer \cite{Phan2022sleeptransformer}, spanning diverse paradigms (convolutional, recurrent, and attention-based) and operating on different input representations (raw signals for U-Sleep and DeepResNet, and spectrograms for SleepTransformer). 
In addition, following the same training protocol, data and splits, we train variants of these architectures using exclusively the EMG modality (Appendix \ref{apd:A_emg}), extending the set of unimodal encoders available for multimodal fusion.

\paragraph{Datasets}
The base models were pre-trained on several PSG datasets available from the National Sleep Research Resource (NSRR) \cite{Zhang2024}, spanning  $\approx 220000$ hours of data from diverse populations.
To prevent data leakage, NAPS models are trained on the hold out sets of the NSRR datasets, and on an independent dataset, unseen by the pre-trained models. Specifically, we use the Bern Sleep-Wake Registry (BSWR) \cite{Aellen2024}.
Overall $9885$ PSG instances, corresponding to $\approx 80000$ hours, are employed for training. 
We ensure a strict separation between data splits by splitting data at the subject level, preventing recordings from the same individual from appearing in different splits. 
Additional out-of-domain (OOD) datasets, which were never seen during the training of either the frozen encoders or NAPS, are used for evaluation: the Danish Centre for Sleep Medicine database (DCSM) \cite{Perslev2021}, the Dreem Open Datasets (DOD-H \& DOD-O) \cite{Guillot2020}, the Sleep-EDF Expanded database (SEDF-SC \& SEDF-ST) \cite{Kemp2000}, and the PhysioNet/CinC 2018
dataset (PHYS) \cite{Ghassemi2018}. 
All recordings are resampled to a sampling rate of 128 Hz and scaled with channel-wise robust scaling.
Further details are provided in Appendix \ref{apd:A_datasets}.
The OOD setup reflects our intended use case: out-of-the-box deployment. The target end-user is not expected to train the meta-model or possess any locally annotated data.

\paragraph{NAPS configurations}
We test the NAPS module under intermediate and late fusion configurations. In the former, NAPS fuses sleep epoch embeddings extracted from base models, whereas in the latter, it processes the vectors of predicted sleep stages probabilities, which can be viewed as a blending approach \cite{Wolpert1992}. To distinguish between the two, we refer to the former as NAPS$_{R}$ (for \textbf{R}epresentations) and the latter as NAPS$_{P}$ (for \textbf{P}redictions).

\paragraph{Role of meta-training data ablation}
To isolate our contributions from the effect of increased data volume, we (i) restrict NAPS$_P$ meta-training to a 1\% subset of the BSWR dataset; (ii) retrain the U-Sleep$_{EEG}$ baseline on its original corpus combined with the full BSWR dataset (Appendix \ref{apd:A_data_ablation}).

\paragraph{Interpretability analysis}
We analyze the attention weights produced by the modality fusion layer in Appendix \ref{app:interpretability}. These weights quantify the predictive importance of each modality across sleep stages. Moreover, we simulate sensor failures during inference (e.g., partial or total signal corruption) to qualitatively evaluate the robustness and adaptive routing capabilities of the fusion layer.

\paragraph{Evaluation}
We report macro F1 (MF1) and per-stage F1 scores. We consider per-recording metrics in the main text and dataset-wise metrics in Appendix \ref{apd:A_dataset_metrics}.
For the DOD datasets, each recording was annotated by five sleep technologists, allowing evaluation against consensus-based scoring (details in Appendix \ref{apd:A_consensus}).
We compare NAPS$_R$ and NAPS$_P$ against the individual base encoders from which they aggregate information, SOMNUS, the corresponding soft-voting ensemble, which has been shown to robustly match or outperform previously proposed methods \cite{DeiRossi2025}, and SleepFM \cite{Thapa2025}, the most prominent PSG foundation model released to date. To ensure a strictly controlled comparison and isolate the effect of data exposure, we fine-tune the SleepFM sleep staging classification head on our exact meta-training splits.
We report architectural ablation studies in Appendix \ref{apd:A_architectural_ablations}. We evaluate two variants: (i) bypassing the tri-axial processing component to directly aggregate marginal representations via the attention-based fusion module; and (ii) replacing the attention-based fusion module with average pooling.
Further comparisons are tackled in the Discussion.

\section{Results}\label{sec:results}

\begin{table*}[t!]
\caption{Per-recording mean (SD) Macro-F1 (MF1) and per-stage F1 (F1$_{stage}$) scores for the best individual unimodal model (soft-voting across channels), the SOMNUS ensemble \cite{DeiRossi2025} (soft-voting
across all channels, modalities and models), NAPS$_P$, and NAPS$_R$. Best results per $<$dataset, metric$>$ are shown in bold. $\ddag$ indicates statistically significant (one-sided paired Wilcoxon signed-rank Bonferroni-Holm corrected, $\alpha<0.05$) MF1 improvement of NAPS over other methods.
}
\label{tab:results}
\begin{center}
\scriptsize
\begin{tabular}{c|l|cccccc}
\hline
\noalign{\vskip 1mm}
\textbf{Dataset} & \textbf{Model} & \textbf{MF1} & \textbf{F1$_{W}$} & \textbf{F1$_{N1}$} & \textbf{F1$_{N2}$} & \textbf{F1$_{N3}$} & \textbf{F1$_{REM}$} \\
\hline
\noalign{\vskip 1mm}
\multirow{4}{*}{BSWR} & DeepResNet$_{EEG}$ & $.692(.126)$ & $.811(.157)$ & $.402(.165)$ & $.799(.148)$ & $.627(.271)$ & $.846(.195)$ \\
& SOMNUS & $.696(.126)$ & $.812(.159)$& $.374(.172)$& $.807(.146)$& $.649(.283)$ & $.862(.183)$ \\ 
& NAPS$_P$ & $.756(.123) \ddag$ & $.842(.142)$ & $.576(.154)$ & $.817(.152)$ & $.705(.266)$ & $.862(.185)$\\
& NAPS$_R$ & $\mathbf{.784(.099) \ddag}$ & $\mathbf{.854(.128)}$ & $\mathbf{.609(.144)}$ & $\mathbf{.837(.109)}$ & $\mathbf{.758(.230)}$ & $\mathbf{.878(.155)}$ \\
\hline\hline
\noalign{\vskip 1mm}
\multirow{5}{*}{DCSM} & DeepResNet$_{EEG}$ & $.797(.086)$ & $.981(.027)$ & $.507(.147)$& $.849(.096)$ & $.779(.207)$ & $.874(.149)$\\
& SOMNUS & $.801(.083)$ & $.983(.023)$& $.497(.150)$& $.858(.096)$& $.778(.206)$ & $.892(.145)$ \\ 
& SleepFM & $.264(.090)$ & $.347(.250)$ & $.081(.092)$ & $.375(.166)$ & $.319(.195)$ & $.202(.195)$ \\
& NAPS$_P$ & $.818(.081) \ddag$ & $\mathbf{.986(.022)}$ & $\mathbf{.564(.139)}$ & $.846(.109)$ & $.806(.191)$ & $.892(.143)$\\
& NAPS$_R$ & $\mathbf{.819(.079)} \ddag$ & $.985(.021)$ & $.547(.149)$ & $\mathbf{.859(.094)}$ & $\mathbf{.813(.185)}$ & $\mathbf{.894(.145)}$\\
\hline
\noalign{\vskip 1mm}
\multirow{5}{*}{DOD-H} & U-Sleep$_{EEG}$ & $.816(.072)$& $.878(.085)$ & $.526(.166)$ & $.907(.051)$ & $.851(.171)$ & $.916(.073)$ \\
& SOMNUS & $.829(.062)$ & $.887(.085)$ & $.542(.159)$ & $\mathbf{.913(.042)}$ & $\mathbf{.870(.162)}$ & $.932(.053)$ \\
& SleepFM & $.662(.132)$ & $.653(.228)$ & $.519(.143)$ & $.782(.108)$ & $.433(.254)$ & $.906(.095)$ \\
& NAPS$_P$ & $\mathbf{.835(.070)}$ & $.878(.099)$ & $\mathbf{.620(.158)}$ & $.901(.049)$ & $.840(.163)$ & $\mathbf{.935(.050)}$ \\
& NAPS$_R$ & $.823(.060)$ & $\mathbf{.890(.079)}$ & $.536(.165)$ & $.903(.044)$ & $.851(.162)$ & $\mathbf{.935(.052)}$ \\
\hline
\noalign{\vskip 1mm}
\multirow{5}{*}{DOD-O} & U-Sleep$_{EEG}$ & $.776(.083)$ & $.906(.076)$ & $.496(.145)$ & $.882(.070)$ & $.696(.264)$ & $.904(.099)$ \\
& SOMNUS & $\mathbf{.790(.083)}$ & $\mathbf{.913(.068)}$ & $.513(.152)$ & $\mathbf{.885(.072)}$ & $\mathbf{.735(.268)}$ & $.912(.078)$ \\
& SleepFM & $.647(.100)$ & $.782(.178)$ & $.444(.152)$ & $.726(.147)$ & $.408(.237)$ & $.871(.136)$ \\
& NAPS$_P$ & $.785(.085)$ & $.879(.103)$ & $\mathbf{.533(.134)}$ & $.864(.079)$ & $.721(.259)$ & $\mathbf{.913(.078)}$ \\
& NAPS$_R$ & $.750(.091)$ & $.894(.087)$ & $.505(.167)$ & $.818(.100)$ & $.636(.260)$ & $.906(.069)$\\
\hline
\noalign{\vskip 1mm}
\multirow{5}{*}{PHYS} & DeepResNet$_{EEG}$ & $.687(.097)$ & $.744(.159)$ & $.358(.153)$ & $.832(.106)$ & $.682(.247)$ & $.837(.173)$\\
& SOMNUS & $.689(.098)$ & $.742(.161)$ & $.338(.158)$ & $\mathbf{.837(.107)}$ & $.697(.251)$ & $.848(.168)$ \\
& SleepFM & $.499(.262)$ & $.593(.266)$ & $.333(.219)$ & $.600(.328)$ & $.394(.345)$ & $.578(.379)$ \\
& NAPS$_P$ & $\mathbf{.744(.095) \ddag}$ & $\mathbf{.793(.148)}$ & $\mathbf{.538(.138)}$ & $.830(.108)$ & $\mathbf{.721(.240)}$ & $.848(.166)$\\
& NAPS$_R$ & $.711(.096) \ddag$ & $.760(.157)$ & $.417(.157)$ & $.833(.105)$ & $.705(.246)$ & $\mathbf{.851(.164)}$ \\
\hline
\noalign{\vskip 1mm}
\multirow{5}{*}{SEDF-SC} & U-Sleep$_{EEG}$ & $.720(.090)$ & $.981(.014)$ & $.342(.130)$ & $.814(.097)$ & $.602(.287)$ & $.845(.114)$ \\
& SOMNUS & $.734(.083)$ & $.982(.018)$ & $.358(.138)$ & $\mathbf{.832(.083)}$ & $\mathbf{.611(.279)}$ & $.870(.094)$ \\
& SleepFM & $.572(.150)$ & $.839(.150)$ & $.345(.147)$ & $.695(.145)$ & $.490(.284)$ & $.484(.297)$ \\
& NAPS$_P$ & $\mathbf{.757(.082) \ddag}$ & $\mathbf{.985(.016)}$ & $\mathbf{.487(.124)}$ & $.822(.087)$ & $.597(.291)$ & $.871(.095)$\\
& NAPS$_R$ & $.739(.083)$ & $\mathbf{.985(.011)}$ & $.454(.120)$ & $.791(.098)$ & $.563(.290)$ & $\mathbf{.876(.084)}$ \\
\hline
\noalign{\vskip 1mm}
\multirow{5}{*}{SEDF-ST} & DeepResNet$_{EEG}$ & $.764(.074)$ & $.814(.105)$ & $.508(.158)$ & $.863(.062)$ & $\mathbf{.746(.232)}$ & $.891(.085)$ \\
& SOMNUS & $.746(.077)$ & $.786(.110)$ & $.452(.143)$ & $\mathbf{.872(.058)}$ & $.716(.233)$ & $\mathbf{.902(.080)}$ \\
& SleepFM & $.480(.192)$ & $.469(.274)$ & $.241(.183)$ & $.673(.190)$ & $.407(.265)$ & $.617(.262)$ \\
& NAPS$_P$ & $\mathbf{.798(.077) \ddag}$ & $\mathbf{.853(.094)}$ & $\mathbf{.618(.153)}$ & $.870(.058)$ & $\mathbf{.746(.232)}$ & $\mathbf{.902(.082)}$ \\
& NAPS$_R$ & $.766(.073) \ddag$ & $.818(.103)$ & $.503(.150)$ & $.868(.056)$ & $.742(.232)$ & $.896(.084)$ \\
\hline
\end{tabular}
\end{center}
\end{table*}

Table~\ref{tab:results} reports the performance of the best unimodal models (determined by MF1) from SLEEPYLAND, their soft-voting ensemble SOMNUS, the SleepFM foundation model, and our proposed attention-based aggregators, NAPS$_P$ and NAPS$_R$. For each instance, aggregation is performed considering all available modalities, channels, and base models. 
Consistent with findings of the original authors, SleepFM struggles to generalize to external cohorts, exhibiting poor zero-shot generalization across OOD datasets. We observe that na\"ive aggregation of base model predictions generally outperforms the best individual base model of the ensemble by a small margin, whereas principled attention-based aggregation proves to be considerably more effective, achieving superior performance in nearly all evaluation scenarios. 
In the in-domain setting (BSWR), we observe a substantial performance boost (MF1 $0.696$ (SOMNUS), $0.756$ (NAPS$_P$), and $0.784$ (NAPS$_R$)). This gain partly reflects the fact that the aggregator benefits from supervised adaptation to the target dataset distribution.
More importantly, across OOD datasets, where no method has access to task-specific labels, attention-based fusion still delivers consistent zero-shot MF1 gains (DCSM: $.801 \to .819$; DOD-H: $.829 \to .835$; PHYS: $.689 \to .744$; SEDF-SC: $.734 \to .757$; SEDF-ST: $.746 \to .798$).
These results indicate that the learned fusion strategies generalize to unseen cohorts and heterogeneous recording conditions.
Notably, the improvements in MF1 are primarily driven by improvements in recognition of the N1 stage, a stage characterized by inherent recognition difficulty and low inter-scorer agreement. 
A distinct trade-off between in-distribution adaptation and out-of-distribution robustness emerges. 
On the BSWR dataset, used for meta-training, intermediate fusion (NAPS$_R$) outperforms late fusion (NAPS$_P$) by a significant margin ($\Delta$MF1 $+2.8\%$).
When the meta-model is exposed to the target distribution, access to high-dimensional marginal representations allows the aggregator to learn domain-specific feature mappings that substantially refine the base models' outputs. Conversely, on the OOD datasets, late fusion typically yields superior generalization compared to intermediate fusion. We attribute this to the stability of the prediction space relative to the feature space. While raw feature distributions are sensitive to acquisition differences inherent in out-of-domain datasets, the output of the base models acts as a normalized, universal interface. 

Appendix \ref{apd:A_additional_experiments} details performance across all modality subsets. Validating the efficacy of dynamic batch sampling, these results demonstrate that NAPS adapts to arbitrary sensor configurations without retraining. We observe a consistent trend where fusing additional modalities yields synergistic gains; remarkably, NAPS retains a significant advantage over other methods even in unimodal settings.

Retraining U-Sleep$_{EEG}$ on the combined NSRR and BSWR corpora doesn't yield improvements compared to its counterpart trained exclusively on the NSRR datasets. This confirms that the superior generalization of NAPS stems from its principled fusion architecture rather than sheer data volume (Appendix \ref{apd:A_data_ablation}).
While integrating either the tri-axial processing or the attention-based fusion module independently surpasses naive ensembling, their combination yields the most robust generalization (Appendix \ref{apd:A_architectural_ablations}). Removing the tri-axial stack causes a clear performance drop. Conversely, replacing attention-based fusion with average pooling incurs only a minor performance penalty but entirely sacrifices the model's transparent predictive routing. Retaining this attention mechanism not only ensures optimal performance but provides crucial interpretability (Appendix \ref{app:interpretability}). An analysis of the fusion weights reveals an intuitive alignment with sleep physiology: NAPS learns to prioritize the EEG modality during deep sleep and reverses this attention during REM, where EOG overtakes EEG as the primary driver. Moreover, when subjected to artificial sensor corruption, this learned mechanism enables robust intra- and inter-modality routing without explicit priors.

\section{Discussion}\label{sec:discussion}

In this work, we introduced NAPS, an attention-based fusion module for multimodal physiological signals. Methodologically, we generalize criss-cross attention \cite{Huang2019} into a tri-axial formulation, yielding a scalable ensembling mechanism that leverages inductive biases proven effective for physiological modeling \cite{Wang2024}. Coupled with a training strategy extending DAT \cite{Malekzadeh2021}, our approach ensures robustness to the flexible sensor configurations inherent in continuous monitoring scenarios.

We validated NAPS on automatic sleep staging using pre-trained supervised unimodal base models, prioritizing zero-shot generalization to unseen cohorts, a prerequisite for scalable deployment that avoids the burden of site-specific retraining. NAPS consistently outperformed both the soft-voting supervised ensemble SOMNUS, its individual constituents, and SleepFM, the most prominent PSG foundation model to date, establishing a new state-of-the-art for zero-shot performance on multiple datasets. While the base models had already convincingly surpassed prior approaches \cite{Vallat2021, Perslev2021}, some performance gaps on OOD datasets such as PHYS and SEDF were observed \cite{DeiRossi2025}. NAPS delivers its most significant gains in these scenarios, suggesting that principled fusion effectively narrows the generalization gap to in-domain performance. 
Our ablations (Appendices \ref{apd:A_architectural_ablations} and \ref{apd:A_data_ablation}) confirm that these gains stem from the architecture's structural inductive biases rather than sheer parameter count or meta-training data volume.
Moreover, unlike average pooling, which obscure the relative contribution of each input, attention-based fusion provides a transparent window into the network's decision-making, revealing that NAPS has learned physiologically sound routing strategies (Appendix \ref{app:interpretability}), such as relying mostly on EEG for deep sleep and on EOG for REM.
Furthermore, our results suggest a trade-off: intermediate fusion proves superior when in-domain adaptation is feasible, whereas late fusion offers greater stability for zero-shot generalization to entirely new cohorts. 

We highlight that, while multimodal self-supervised learning offers a promising avenue for developing foundation models adaptable to diverse health-related tasks \cite{Thapa2025}, such approaches currently lag behind supervised alternatives. This gap is particularly acute in zero-shot generalization to new cohorts. 
For instance, recent self-supervised efforts report a dataset-wise MF1 of $0.718$ on PHYS \cite{Fang2024} after fine tuning, over 6\% lower than the OOD performance achieved by NAPS$_P$ ($0.781$, Appendix \ref{apd:A_dataset_metrics}), 
without any cohort-specific tuning. 
This trend extends to SleepFM \cite{Thapa2025}, which trails current supervised benchmarks \cite{Perslev2021, DeiRossi2025} by a small margin on in-domain datasets, but exhibits dramatic degradation in out-of-domain zero-shot settings. As explicitly acknowledged by its original authors, the model struggles to generalize to external validation cohorts, a vulnerability our results starkly confirm. Evaluated zero-shot across all six unseen external cohorts, SleepFM exhibits severe and widespread degradation, falling considerably short of both the supervised SOMNUS ensemble and our NAPS models across every evaluated scenario.
Similarly, \cite{Fox2025} reports severe degradation (MF1 0.567) when generalizing to unseen cohorts like MESA. 
As such, supervised approaches currently offer significantly more robust out-of-the-box generalization for the task of automatic sleep staging.

We foresee several avenues for extension; future implementations of PSG modeling could incorporate diverse segment-level representations, such as semantic embeddings from clinical text or symbolic descriptors of traditional waveform events like spindles and K-complexes. 
Beyond PSG, the methodological principles underlying NAPS extend to any multivariate time-series task characterized by decomposable heterogeneity, as the architecture is explicitly engineered to navigate complexity arising from concurrent modalities, variable channel counts, and diverse representational views. 
By coupling the efficiency of tri-axial attention with the robustness of dynamic batching, this methodology is uniquely suited for the adaptive fusion of heterogeneous sensor streams, particularly in high-density or long-context settings. 
For instance, NAPS could be leveraged in advanced wearable systems that integrate sensors into digital textiles \cite{Wicaksono2020, Ryu2021},
or within emerging Internet of Bodies frameworks \cite{Matwyshyn2019, Celik2021}, where adaptive fusion could facilitate the integration of diverse sensor data into a unified physiological representation.

Finally, while this study showcased NAPS as a modular aggregator for frozen representations or predictions, the architecture is differentiable and thus viable as a fusion module within fully end-to-end networks. This setting, however, introduces optimization challenges, most notably the phenomenon of modality competition \cite{Huang2022}.
While our proposed dynamic batching strategy mitigates this risk by introducing partial sensor unavailability, it remains an open question whether this stochasticity alone is sufficient to prevent unimodal dominance. Future research should investigate the interplay between dynamic sampling, modality dropout, and explicit regularization strategies, such as auxiliary losses proposed by \cite{Kontras2024b, Kontras2025}, to ensure robust representation learning across heterogeneous sensor streams.


\bibliography{biblio}
\bibliographystyle{icml2026}


\clearpage

\appendix

\section{Appendix}

\subsection{Empirical Validation of Tri-Axial Attention Efficiency}\label{apd:A_complexity}
\begin{figure}[h]
    \centering
    \includegraphics[width=\textwidth]{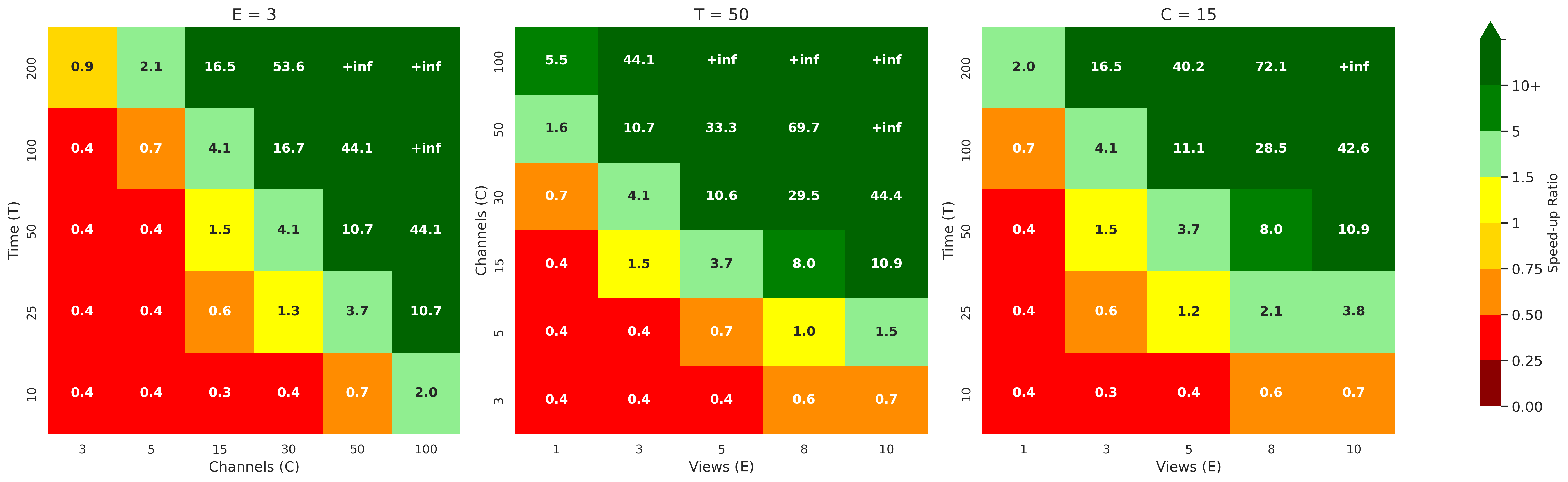}
    \caption{
        Wall-Clock Speed-up Ratio ($T_{FSA} / T_{TAA}$) of TAA over FSA.
        The heatmaps illustrate the observed ratio of execution time for FSA to TAA across various combinations of Time ($T$), Channels ($C$), and Views ($E$), fixing one dimension at a time. Regions colored Red/Orange (Ratio $< 1.0$) indicate that the TAA is slower due to computational overhead for low-dimensional inputs. The Yellow/Green regions (Ratio $\ge 1.0$) show TAA's superior scalability, achieving speed-ups of ${10\times}$ or more (Dark Green) as individual dimensions increase. Dark green regions in the upper right marked with \emph{+inf} indicate instances where FSA failed due to Out-Of-Memory (OOM) errors while TAA remained feasible and fast.
    }
    \label{fig:compl_ratio}
\end{figure}

\begin{figure}[h]
    \centering
    \includegraphics[width=\textwidth]{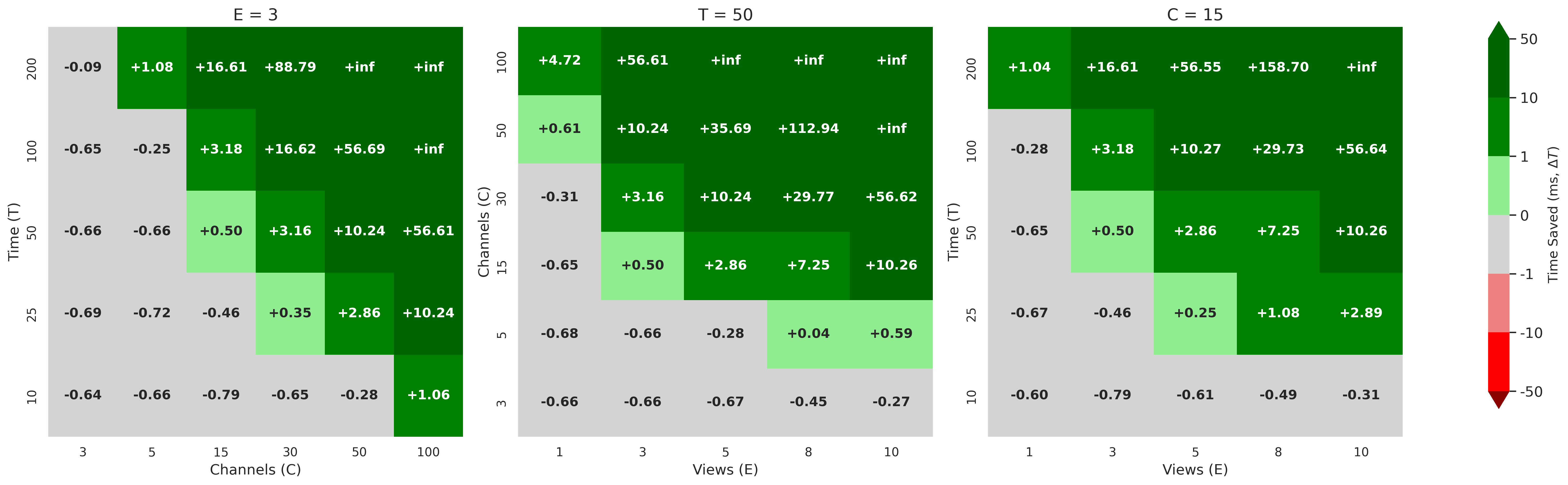}
    \caption{
        Wall-Clock Time Saved ($\Delta T = T_{FSA} - T_{TAA}$) by TAA over FSA (in milliseconds), across various combinations of Time ($T$), Channels ($C$), and Views ($E$), fixing one dimension at a time.
        Red/Pink regions ($\Delta T < 0$) show that TAA is slower for small input sizes, incurring a time penalty (overhead). Green regions ($\Delta T > 0$) highlight the significant time saved, demonstrating TAA's massive advantage for large inputs. Dark green regions in the upper right marked with \emph{+inf} indicate instances where FSA failed due to Out-Of-Memory (OOM) errors while TAA remained feasible and fast.
    }
    \label{fig:compl_delta}
\end{figure}

The Tri-Axial Attention (TAA) mechanism is engineered to provide a substantial theoretical reduction in computational complexity and memory requirement, with respect to the $T$, $C$, and $E$ dimensions, compared to the original Full Self-Attention (FSA) mechanism of \cite{Vaswani2017}, with the latter having a complexity of $\mathcal{O}$$(T \cdot C \cdot E)^2$ and the former of $\mathcal{O}$$(T^2 C E + T C^2 E + T C E^2)$.
We complement the theoretical advantage with an empirical analysis of wall-clock timing of FSA and TAA on a single GPU, for tensors of varying $T, C, E$ dimensions. We measure the Speed-up Ratio ($\frac{T_{FSA}}{T_{TAA}}$) (Figure \ref{fig:compl_ratio}) and the Time Saved ($\Delta T = T_{FSA} - T_{TAA}$) in milliseconds (Figure \ref{fig:compl_delta}). $d_{model}$ is fixed to 36 in this analysis. Results reported are averaged over 50 runs for every configuration.

We observe that when the total sequence length $L = T \cdot C \cdot E$ is small, the TAA mechanism is slower than FSA, resulting in a Speed-up Ratio $< 1.0$ (Red and Orange regions) and a negative Time Saved ($\Delta T < 0$). This loss in efficiency is attributed to the computational overhead associated with TAA's factorized design, including multiple tensor re-arrangements, concatenations, and normalizations. For small tensors, this fixed overhead outweighs the reduced theoretical complexity. We note, however, that the absolute wall-clock time in such cases is still minimal, as indicated by the low magnitude of $\Delta T$.
As the dimensions $T, C,$ and $E$ increase, wall-clock time of FSA rapidly grows, allowing the TAA's more favorable complexity to dominate. 
The break-even point (Speed-up Ratio $= 1.0$, $\Delta T \approx 0$ ms) is crossed when $L$ reaches a critical threshold of approximately 2000 segments. Beyond this point, TAA delivers increasing speed-ups, reaching ratios over $10\times$. The empirical results confirm TAA as a far more scalable attention mechanism in high-density or long-sequence sensing scenarios.
For the largest configurations tested, FSA becomes infeasible, resulting in an out-of-memory (OOM) error. This behavior is a direct consequence of the $\mathcal{O}((TCE)^2)$ memory requirement for storing the dense attention score matrix $Q K^T$. By contrast, the Tri-Axial Attention only requires memory proportional to the sum of the squared individual axis lengths scaled by the remaining dimensions, allowing it to remain feasible and fast even in settings where FSA is functionally impossible.
Furthermore, we point out that, while hardware-aware optimizations like Flash Attention \cite{Dao2022} reduce the memory footprint of exact attention, they do not alter its quadratic computational complexity. As sequence lengths grow, the compute cost eventually becomes prohibitive regardless of memory efficiency. Tri-axial attention provides a strictly more scalable complexity class for high-dimensional sensor arrays, and could itself be implemented using IO-aware optimizations to combine algorithmic efficiency with hardware acceleration. 

\subsection{Dynamic Batch Sampling}\label{apd:A_batching}

The following algorithm determines the dimensions of a single batch.

\begin{algorithm}[H]
\label{alg:batching}
\KwIn{$M^{\max}$, $\{C_{m_k}^{\max}\}_{k=1}^{M_{\max}}$, $\{B_{m_k}^{\max}\}_{k=1}^{M_{\max}}$}
\KwOut{\mbox{Batch dimensions} \; $\{T, M, \{C_{m_k}\}_{k=1}^M, \{B_{m_k}\}_{k=1}^M\}$}

\vspace{1.5\baselineskip}

$T \sim \mathcal{U}\{20,60\}$ \tcp*[r]{sequence length}
$M \sim \mathcal{U}\{1, M_{\max}\}$ \tcp*[r]{modalities} 

\vspace{0.5\baselineskip}

\For{$k \gets 1$ \KwTo $M$}{
    $C_{m_k} \sim \mathcal{U}\{1, C_{m_k}^{\max}\}$ \tcp*[r]{channels}
    $B_{m_k} \sim \mathcal{U}\{1, B_{m_k}^{\max}\}$ \tcp*[r]{base models}
}
\end{algorithm}

Based on the returned dimensions, the specific modalities, channels, and base models are then uniformly sampled from the available options within the observations belonging to the batch.

\subsection{Implementation Details}\label{apd:A_hparams}

This section details the model architecture specifications and the training protocol.

We perform hyperparameter tuning by considering variations across several key hyperparameters: learning rate $\eta \in \{0.001, 0.0005, 0.0001\}$, model dimension $d_{\text{model}} \in \{36, 72, 144, 288\}$, attention heads $h \in \{6, 9, 18\}$ ($h/3$ heads per pathway in the tri-axial mechanism), encoder layers $L \in \{3, 6, 9\}$, and classifier dropout probability  $p_{head} \in \{0.00, 0.25\}$. We also treat the weight-sharing strategy across modalities within the tri-axial encoders as a tunable hyperparameter. This structural choice determines the placement of the modality embedding $\mathbf{ME}_{(m_k)}$: in the case of a single shared transformer, the embedding is added to the input to provide a distinct tag that allows shared parameters to contextualize diverse inputs; conversely, when using independent modality-specific encoders, the embedding is added to the tensor after the tri-axial processing.
To constrain the search space, consistent with prior work \cite{Phan2022sleeptransformer}, we fix the attention inner dimension to $d_A = d_{\text{model}}/2$ and the feed-forward dimension to $d_{\text{ff}} = 4 \cdot d_{\text{model}}$ \cite{Vaswani2017}. We employ the GeLU activation function \cite{Hendrycks2016} and apply dropout throughout the transformer components with a rate of $p = 0.1$.

We report the configurations selected as optimal based on validation MF1 scores, and used for all reported results.
For both NAPS$_P$ and NAPS$R$, the best performance is achieved using independent modality-specific encoders (no weight sharing) with a model dimension of $d{_\text{model}} = 72$, $h=9$ attention heads, and a depth of $L=6$ layers. NAPS$_P$ utilizes a learning rate of $\eta = 1 \times 10^{-3}$, while NAPS$_R$ requires a slightly lower rate of $\eta = 5 \times 10^{-4}$.
This additional parameter footprint (1.5M trainable parameters), along with the associated inference computational cost, is negligible when compared to the overhead of the pre-trained base models used for feature extraction.

During training, each batch includes $B = 8$ recordings. For every recording, we randomly sample $K = 4$ sequences of the same random length. Gradients are accumulated over $G = 4$ forward–backward passes, resulting in an effective batch size of $G \cdot B \cdot K = 128$ sequences per optimization step. Given this stochastic sampling procedure, we define a \emph{training epoch} not as a full pass over the dataset, but as a fixed duration of 200 batches. 
Optimization is performed using AdamW \cite{Loshchilov2017} for a maximum of 300 epochs, with early stopping triggered if the validation macro-F1 score does not improve for 30 consecutive epochs. We employ a composite learning rate schedule: a linear warmup increases the learning rate from $0.1\eta$ to $\eta$ over the first 10 epochs, followed by a cosine annealing schedule \cite{Loshchilov2017} that decays the learning rate to a minimum of $10^{-8}$ over the remaining epochs.

For the reported results, we run inference on one recording at a time, utilizing all available modalities, channels, and base models. Recordings are processed in non-overlapping segments of length $T=35$ sleep epochs, consistent with the windowing used in \cite{Perslev2021, DeiRossi2025}. While averaging predictions over overlapping sliding windows can yield marginal performance gains, we adhere to non-overlapping inference to minimize computational overhead.

NAPS and the experimental pipeline are implemented in PyTorch 2.8.0 and Einops 0.8.1 \cite{Rogozhnikov2022}. Hyperparameter tuning is managed via Hydra \cite{Yadan2019}, and training is performed on a single NVIDIA L40S GPU.

\subsection{Datasets}\label{apd:A_datasets}
This section provides a summary description of all datasets used in our experiments, with appropriate references and links for the detailed reports.
\newline

\subsubsection{Training Datasets}

We consider during the meta-training phase of NAPS models the BSWR dataset, described first, and NSRR datasets \cite{Zhang2018, Zhang2024}, more specifically their hold-out sets as defined in SLEEPYLAND \cite{DeiRossi2025}, to avoid any overlapping data in the training of base predictors and meta-models.

\paragraph{BSWR}
The \emph{Bern Sleep-Wake Registry} (BSWR) \cite{Aellen2024} is a private dataset which comprises a total of 8,410 PSG recordings ($\approx 67'000$ hours) from patients aged 0–91 years, collected during routine clinical practice. This dataset uniquely covers the full spectrum of sleep-wake disorders, including cases with multiple comorbidities and non-sleep-related conditions.  Only a small fraction of participants (\(<1\%\)) are healthy controls, while the majority are patients diagnosed with one or more sleep disorders or cases with uncertain diagnoses. Among the recorded disorders, sleep-related breathing disturbances are the most prevalent, followed by central hypersomnolence disorders, insomnia, parasomnias, and sleep-related movement disorders. A smaller subset of patients present circadian rhythm disorders or isolated symptoms without a definitive clinical classification. 
We consider EEG signals (F4-M1, F3-M2, C4-M1, C3-M2, O2-M1, O1-M2) and EOG signals (E2-M1, E1-M2), sampled at 200 Hz. All recordings are manually annotated by certified sleep experts following the American Academy of Sleep Medicine (AASM) guidelines \cite{Berry2017}. The dataset is partitioned into training, validation, and test, with splits performed by considering subject identifiers, using a 90/5/5 ratio.\\

\textbf{Ethical approval :} The secondary usage of the BSWR dataset was approved by the ethics committee, 
ensuring compliance with the Human Research Act (HRA) and Ordinance on Human Research with the Exception of Clinical Trials (HRO). All methods were carried out in accordance with relevant guidelines and regulations. Written informed consent was obtained from participants as of the introduction of the general consent process at Inselspital in 2015. Data were maintained with confidentiality throughout the study.

\paragraph{NSRR Datasets}
The National Sleep Research Resource (NSRR) is an NHLBI-supported data repository designed to promote open sharing of large-scale sleep research data \cite{Zhang2018, Zhang2024}. Established in 2014, NSRR provides access to polysomnography, actigraphy, and questionnaire-based datasets collected from diverse cohorts and clinical studies. By enabling secondary analyses, algorithm development, and signal processing research, NSRR aims to advance sleep and circadian science. The repository currently hosts tens of thousands of high-quality sleep records. More info: \url{https://sleepdata.org/pages/about}.

\textbf{ABC}.
The Apnea, Bariatric surgery, and CPAP study includes 132 recordings from 49 patients with severe OSA and morbid obesity (BMI 35–45) \cite{Bakker2018}. EEG (F3-M2, F4-M1, C3-M2, C4-M1, O1-M1, O2-M2), EOG (E1-M2, E2-M1), and EMG (center chin, left submentalis, right submentalis) were acquired at 256 Hz, band-pass filtered, and scored according to AASM criteria. More info: \url{https://clinicaltrials.gov/ct2/show/NCT01187771}. We consider 35 recordings from the hold out set of SLEEPYLAND.

\textbf{APOE}. 
The Sleep Disordered Breathing, apolipoprotein E, and
Lipid Metabolism dataset is a study investigating genetic associations with sleep-disordered breathing, comprising 712 PSGs from untreated participants stratified by ApoE genotype \cite{Moore2014}. EEG (C3-M2, C4-M1, O2-M1, O1-M2, C3-M1, C4-M2, O2-M2, O1-M1, F1-M2, F2-C4, F2-T4, FP1-C3, FP1-C3, FP2-C4, Fz-M1, Fz-M2, T3-O1 T4-O2) and EOG
(ROC-M1, LOC-M2) were recorded at 256 Hz, while EMG (Chin-EMG, Chin-L, Chin-R, Chin-Ctr) was recorded at 512 Hz, and scored according to AASM criteria. More info: \url{https://doi.org/10.25822/6ssj-2157}. We consider 150 recordings from the hold out set of SLEEPYLAND.  

\textbf{APPLES}. 
The Apnea Positive Pressure Long-term Efficacy
Study is a multi-center randomized clinical trial on positive airway pressure for OSA, with 1094 PSGs \cite{Quan2011}. EEG (C3-M2, C4-M1, O2-M1, O1-M2), EOG (ROC-M1, LOC-M2), and EMG (submentalis) signals are recorded at 128 Hz, initially scored according to Rechtschaffen and Kales scoring rules (R\&K) and then re-aligned to AASM \cite{Moser2009}. More info: \url{https://clinicaltrials.gov/study/NCT00051363?tab=results}. We consider 150 recordings from the hold out set of SLEEPYLAND. 

\textbf{CCSHS}. 
The Cleveland Children’s Sleep and Health Study includes 515 PSGs \cite{Rosen2003} from three different cohorts in Cleveland, Ohio, USA. EEG (C3-A2, C4-A1), EOG (ROC-A1, LOC-A2), EMG (center chin, left submentalis, right submentalis) were recorded at 128 Hz, and manually scored according to AASM rules. More info: \url{https://doi.org/10.25822/cg2n-4y91}. We consider 128 recordings from the hold out set of SLEEPYLAND.

\textbf{CFS}. 
The Cleveland Family Study is a family-based study on OSA \cite{Redline1995}. SLEEPYLAND used 730 PSGs from 144 families, with splits respecting family membership. EEG (C3-A2, C4-A1) and EOG (ROC-A1, LOC-A2) signals were recorded at 128 Hz, while EMG (center chin, left submentalis, right submentalis) was recorded at 256 Hz, and scored according to AASM rules. More info: \url{https://doi.org/10.25822/jmyx-mz90}. We consider 185 recordings from the hold out set of SLEEPYLAND.

\textbf{CHAT}. 
The Childhood Adenotonsillectomy Trial includes 1638 PSGs from 1232 children (age range: 5–10) post-adenotonsillectomy-surgery with mild-to-moderate OSA across six U.S. centers \cite{Marcus2013}. EEG (F4-M1, F3-M2, C4-M1, C3-M2, O2-M1,
O1-M2, T4-M1, T3-M2), EOG (E2-M1, E1-M2), and EMG (CChin, LChin, RChin, LChin-Rchin) signals were recorded at $\geq 200$ Hz, and scored according to AASM rules. More info: \url{https://clinicaltrials.gov/study/NCT00560859}. We consider 199 recordings from the hold out set of SLEEPYLAND.

\textbf{HOMEPAP}. 
The Home Positive Airway Pressure dataset is a multi-site U.S. study on home PAP therapy \cite{Rosen2012}, with 246 PSGs considered in SLEEPYLAND. We consider the EEG (F4-M1, F3-M2, C4-M1, C3-M2, O2-M1, O1-M2, T4-M1, T3-M2), EOG (E2-M1, E1-M2), and EMG (center chin, left submentalis, right submentalis) signals originally recorded at 200 Hz, and scored according to AASM scoring rules. More info: \url{https://clinicaltrials.gov/ct2/show/NCT00642486}. We consider 62 recordings from the hold out set of SLEEPYLAND.  

\textbf{MESA}. 
The Multi-Ethnic Study of Atherosclerosis includes 2056 PSGs from adults aged 45–84 across four ethnic groups \cite{Chen2015}. EEG (Fz-Cz, C4-M1, CzOz), EOG (E2-Fpz, E1-Fpz), and EMG (Chin) signals were recorded at 256Hz, low-pass filtered at 100 Hz, and scored by sleep experts according to the AASM rules. More info: \url{ https://doi.org/10.25822/n7hq-c406}. We consider 150 recordings from the hold out set of SLEEPYLAND.   

\textbf{MNC}.
The Mignot Nature Communications dataset comprises $\approx 1000$ PSGs used in \cite{Stephansen2018}. Sub-cohorts include CNC (78 PSGs, of which we consider 20 for NAPS training), DHC (83 PSGs, of which we consider 22 for NAPS training), and SSC (767 PSGs, of which we consider 150 for NAPS training). EEG (C3-M2, C3, C4-M1, C4, Cz,
F3-M2, F3, F4-M1, F4, O1-M2, O1, O2-M1, O2), EOG (E1-M2
E1 E2-M1 E2), and EMG (CChin, LChin, Chin) signals were recorded at 128Hz, and manually scored by sleep experts according to the AASM rules. More info: \url{https://doi.org/10.25822/00tc-zz78}.

\textbf{MROS}.
A subset of the Osteoporotic Fractures in Men study \cite{Blackwell2011}, with 3930 PSGs from older men ($>65$ years), most affected by sleep disorders. EEG (C4-A1, C3-A2), EOG (ROC-A1, LOC-A2), and EMG (LChin, RChin, LChin-RChin) signals were recorded at 256 Hz, and scored according to AASM rules. More info: \url{https://doi.org/10.25822/kc27-0425}. We consider 195 recordings from the hold out set of SLEEPYLAND. 

\textbf{MSP}.
The Maternal Sleep in Pregnancy dataset \cite{Dipietro2021} is comprised of 105 overnight PSGs from women at week 36 of pregnancy, without previously
identified sleep disorders. EEG (C3-M2, C4-M1, F3-M2, F4-M1, O1-M2, O2-M1), EOG (LOC, ROC), and EMG (Chin) signals were recorded at 256Hz and scored according to the AASM manual. More info: \url{https://sleepdata.org/datasets/msp}. We consider 27 recordings from the hold out set of SLEEPYLAND.

\textbf{NCHSDB}. 
The Nationwide Children’s Hospital Sleep DataBank consists of 3950 pediatric PSGs (age range: 0–18) \cite{Lee2022}. EEG (FP1, FP2, FZ, CZ, PZ, OZ, FPZ, P3-M2, P4-M1, F3-M2, F4-M1, F4-M2, C3-M2, C4-M1, C4-M2, T3-M2, T4-M1, O1-M2, O2-M1, F4, O1, O2), EOG (E1-M2, E2-M1, E1, E2), and EMG (Chin1, Chin2) signals were recorded at 256 Hz for most recordings. Recordings were manually scored following AASM criteria. More info: \url{https://sleepdata.org/datasets/nchsdb}. We consider 161 recordings from the hold out set of SLEEPYLAND.

\textbf{SHHS}. 
The Sleep Heart Health Study is a large dataset that comprises 8444 PSGs from 5797 adults ($\geq$ 40 years), most of which suffering from sleep disorders, across two visits \cite{Quan1997}. EEG (C3-A2, C4-A1), EOG (ROC-A1, LOC-A2), and EMG (Chin) signals were recorded at sampling frequencies of 125 Hz, 50 Hz, and 125 Hz, respectively. Recordings were initially R\&K scored and subsequently re-aligned to AASM scoring rules. More info: \url{https://clinicaltrials.gov/ct2/show/NCT00005275}. We consider 221 recordings from the hold out set of SLEEPYLAND.  

\textbf{SOF}. 
We consider a subset of the Study of Osteoporotic Fractures \cite{Spira2008}, with 453 PSGs from older women. We consider EEG (C3-A2, C4-A1), EOG (ROC-A1, LOC-A2), and EMG (LChin, RChin) signals which were recorded at 128 Hz, initially R\&K scored, and re-aligned with AASM criteria. More info: \url{https://doi.org/10.25822/e1cf-rx65}. We consider 114 recordings from the hold out set of SLEEPYLAND.  

\textbf{WSC}. 
Wisconsin Sleep Cohort is an ongoing longitudinal study investigating the causes, consequences, and natural history of sleep disorders; SLEEPYLAND considers 2569 in-laboratory PSGs across four visits \cite{Young2009}. EEG (F3-M1, F3-M2, F3-AVG, F4-M1, F4-M2, F4-AVG, Fz-M1, Fz-M2, Fz-AVG, Cz-M1, Cz-M2, Cz-AVG, C3-M1, C3-M2, C3-AVG, C4-M1, C4-M2, C4-AVG, Pz-M1, Pz-M2, Pz-AVG, Pz-Cz, O1-M1, O1-M2, O1-AVG, O2-M1, O2-M2, O2-AVG), EOG (E1, E2), and EMG (chin, cchin-l) are included, recorded at sampling rates of either 100 Hz or 200 Hz, depending on the system. Recordings are scored by sleep experts according to AASM criteria. More info: \url{https://sleepdata.org/datasets/wsc}. We consider 347 recordings from the hold out set of SLEEPYLAND. 

\subsubsection{Evaluation Datasets}

The following datasets are used exclusively in inference; neither SLEEPYLAND base encoders   nor NAPS models were trained on recordings from these datasets, enabling evaluation of zero-shot performance.

\begin{table}[htbp]
\centering
\caption{Summary statistics of evaluation datasets, reporting the number of PSG recordings, average participant age (mean $\pm$ standard deviation), and gender distribution, where available.}
\label{tab:db_overview}
\renewcommand{\arraystretch}{1.15} 
\setlength{\tabcolsep}{5pt}       
\begin{tabular}{lccc}
\toprule
\textbf{Dataset} & \textbf{\# PSGs} & \textbf{Age (years)} & \textbf{F/M (\%)} \\
\midrule
DCSM      & $255$  & $-$             & $-$ \\
DOD-H     & $25$   & $35.3 \pm 7.5$  & $24/76$ \\
DOD-O     & $55$   & $45.6 \pm 16.5$ & $36/64$ \\
PHYS      & $994$  & $55.2 \pm 14.3$ & $33/67$ \\
SEDF-SC   & $153$  & $58.8 \pm 22.0$ & $53/47$ \\
SEDF-ST   & $44$   & $40.2 \pm 17.7$ & $68/32$ \\
\bottomrule
\end{tabular}
\end{table}

\textbf{DOD}. 
The \emph{Dreem Open Datasets} consist of two subsets, \textbf{DOD-H} and \textbf{DOD-O} \cite{Guillot2020}. 
DOD-H includes 25 recordings from healthy adults (19 males, 6 females) aged 18–65 years, collected at the Fatigue and Vigilance Unit of the French Armed Forces Biomedical Research Institute (IRBA), Bretigny-Sur-Orge, France. We use EEG channels (C3-M2, C4-M1, F3-F4, F3-M2, F3-O1, F4-O2, O1-M2, O2-M1) along with left and right EOG signals, and a single EMG derivation, sampled at 512 Hz.  
DOD-O contains 55 PSG recordings from patients diagnosed with obstructive sleep apnea (35 males, 20 females) aged 39–62 years, collected at the Stanford Sleep Medicine Center. EEG signals include (C3-M2, C4-M1, F4-M1, F3-F4, F3-M2, F3-O1, F4-O2, FP1-F3, FP1-M2, FP1-O1, FP2-F4, FP2-M1, FP2-O2), left/right EOG, and EMG. Recordings are sampled at 250 Hz.  
All signals undergo preprocessing: a Butterworth IIR band-pass filter \([0.4, 18]\) Hz is applied, recordings are resampled to 100 Hz, clipped, and scaled by dividing by 500 to mitigate extreme amplitude variations. Sleep stages are scored by five physicians across three independent centers using AASM guidelines. \\

\textbf{DCSM}. The \emph{Danish Centre for Sleep Medicine} (DCSM) dataset \cite{Perslev2021} consists of 255 PSG recordings from patients referred for suspected or nonspecific sleep-related disorders. No demographic metadata is provided. We include EEG (F4-M1, F3-M2, C4-M1, C3-M2, O2-M1, O1-M2, T4-M1, T3-M2), EOG (E2-M1, E1-M2), and EMG channels sampled at 256 Hz. A band-pass filter between 0.3 Hz and 70 Hz is applied. All recordings are scored manually by certified clinicians according to AASM criteria. Additional dataset details are available at \url{https://erda.ku.dk/public/archives/db553715ecbe1f3ac66c1dc569826eef/published-archive.html}.\\

\textbf{SEDF}. The \emph{Sleep-EDF Expanded} dataset \cite{Goldberger2000, Kemp2000} consists of two subsets, \textbf{SEDF-SC} and \textbf{SEDF-ST}. SEDF-SC (Sleep Cassette) is comprised of 153 PSG recordings from 78 healthy participants aged 25–101 years.  SEDF-ST (Sleep Telemetry) includes 44 recordings from 22 subjects.  
For our experiments, we use EEG (Fpz-Cz, Pz-Oz), and EOG (ROC-LOC) sampled at 100 Hz. For SEDF-ST we also consider submental-EMG (100 Hz), while for SEDF-SC it isn't available with sufficient sampling rate (1 Hz). Original annotations, scored according to R\&K criteria, were re-aligned to match the AASM scoring standard. Additional details are available at \url{https://doi.org/10.13026/C2C30J}.

\textbf{PHYS}. The dataset from the \emph{PhysioNet/Computing in Cardiology Challenge 2018} \cite{Goldberger2000, Ghassemi2018} includes 1,985 overnight PSG recordings, of which we use 994 labeled sessions in our experiments. EEG channels (F4-M1, F3-M2, C4-M1, C3-M2, O2-M1, O1-M2), one EOG channel (E1-M2), and one EMG channel are considered. Recordings are sampled at 200 Hz and manually annotated following AASM guidelines. Full documentation can be found at \url{https://physionet.org/content/challenge-2018/1.0.0/}.

\subsection{Evaluation Against Medical Consensus}\label{apd:A_consensus}

We adopt the multi-annotator evaluation framework introduced in \cite{Guillot2020} for DOD datasets. Each recording in DOD is independently annotated by $S=5$ experienced sleep technologists, allowing model performance to be evaluated relative to both individual scorers and collective consensus.

Given $S$ scorers, let $y_s^t \in \{0,1,2,3,4\}$ denote the label assigned by scorer $s$ to epoch $t$ and $\hat{y}_s^t \in \{0,1\}^5$ its one-hot encoding. For scorer $s$, we define the agreement of the remaining scorers at epoch $t$ as:
\begin{equation}
    \hat{z}_s^t = \frac{\sum_{i \neq s} \hat{y}_i^t}{\max \left( \sum_{i \neq s} \hat{y}_i^t \right)}.
\end{equation}
The \emph{soft-agreement} of scorer $s$ over a recording is:
\begin{equation}
    \text{Soft-Agreement}_s = \frac{1}{T} \sum_{t=1}^T \hat{z}_s^t[y_s^t],
\end{equation}
which measures how often the scorer aligns with the collective judgment, weighted by inter-scorer agreement. Reliable scorers are defined as those with the highest soft-agreement scores for a given recording.
The discrete consensus hypnogram is obtained by majority voting across scorers, with ties resolved using the most reliable scorer. 

\subsection{Dataset-wise Performance Metrics}\label{apd:A_dataset_metrics}

\begin{table*}[hbt!]
\caption{Dataset-wise Macro-F1 (MF1) and per-stage F1 (F1$_{stage}$) scores for the best individual unimodal model (soft-voting across channels), the SOMNUS ensemble \cite{DeiRossi2025} (soft-voting across all channels, modalities and models), SleepFM \cite{Thapa2025}, NAPS$_P$, and NAPS$_R$. Best results per dataset and metric are shown in bold.}
\label{tab:results_global}
\begin{center}
\scriptsize
\begin{tabular}{c|l|cccccc}
\hline
\noalign{\vskip 1mm}
\textbf{Dataset} & \textbf{Model} & \textbf{MF1} & \textbf{F1$_{W}$} & \textbf{F1$_{N1}$} & \textbf{F1$_{N2}$} & \textbf{F1$_{N3}$} & \textbf{F1$_{REM}$} \\
\hline
\noalign{\vskip 1mm}
\multirow{4}{*}{BSWR} & SleepTransformer$_{EEG}$ & $.739$ & $.840$ & $.378$ & $.826$ & $.764$ & $.886$ \\
& SOMNUS & $.742$ & $.849$ & $.373$ & $.832$ & $.755$ & $.900$ \\ 
& NAPS$_P$ & $.802$ & $.877$ & $.590$ & $.844$ & $.800$ & $.901$ \\
& NAPS$_R$ & $\mathbf{.820}$ & $\mathbf{.888}$ & $\mathbf{.622}$ & $\mathbf{.852}$ & $\mathbf{.830}$ & $\mathbf{.906}$ \\
\hline\hline
\noalign{\vskip 1mm}
\multirow{5}{*}{DCSM} & DeepResNet$_{EEG}$ & $.814$ & $.982$ & $.515$ & $.857$ & $.827$ & $.887$ \\
& SOMNUS & $.816$ & $.984$ & $.502$ & $.866$ & $.821$ & $.905$ \\ 
& SleepFM & $.326$ & $.512$ & $.116$ & $.412$ & $.364$ & $.216$ \\
& NAPS$_P$ & $.832$ & $\mathbf{.987}$ & $\mathbf{.574}$ & $.855$ & $.842$ & $.904$ \\
& NAPS$_R$ & $\mathbf{.835}$ & $.986$ & $.565$ & $\mathbf{.867}$ & $\mathbf{.849}$ & $\mathbf{.907}$ \\
\hline
\noalign{\vskip 1mm}
\multirow{5}{*}{DOD-H} & USleep$_{EEG}$ & $.834$ & $.902$ & $.565$ & $.908$ & $.869$ & $.926$ \\
& SOMNUS & $.847$ & $\mathbf{.918}$ & $.579$ & $\mathbf{.915}$ & $\mathbf{.882}$ & $.939$ \\ 
& SleepFM & $.667$ & $.660$ & $.549$ & $.789$ & $.426$ & $.912$ \\
& NAPS$_P$ & $\mathbf{.854}$ & $.914$ & $\mathbf{.662}$ & $.903$ & $.850$ & $.941$ \\
& NAPS$_R$ & $.841$ & $\mathbf{.918}$ & $.577$ & $.905$ & $.862$ & $\mathbf{.943}$ \\
\hline
\noalign{\vskip 1mm}
\multirow{5}{*}{DOD-O} & USleep$_{EEG}$ & $.795$ & $.915$ & $.495$ & $.885$ & $.768$ & $.912$ \\
& SOMNUS & $\mathbf{.808}$ & $\mathbf{.918}$ & $.513$ & $\mathbf{.888}$ & $\mathbf{.804}$ & $.918$ \\ 
& SleepFM & $.655$ & $.802$ & $.445$ & $.735$ & $.410$ & $.884$ \\
& NAPS$_P$ & $.795$ & $.890$ & $\mathbf{.524}$ & $.869$ & $.768$ & $\mathbf{.920}$ \\
& NAPS$_R$ & $.761$ & $.899$ & $.514$ & $.821$ & $.664$ & $.909$ \\
\hline
\noalign{\vskip 1mm}
\multirow{5}{*}{PHYS} & DeepResNet$_{EEG}$ & $.724$ & $.788$ & $.360$ & $.845$ & $.760$ & $.869$ \\
& SOMNUS & $.726$ & $.788$ & $.336$ & $\mathbf{.850}$ & $.774$ & $.880$ \\ 
& SleepFM & $.566$ & $.522$ & $.375$ & $.689$ & $.556$ & $.686$ \\
& NAPS$_P$ & $\mathbf{.781}$ & $\mathbf{.836}$ & $\mathbf{.553}$ & $.844$ & $\mathbf{.790}$ & $.879$ \\
& NAPS$_R$ & $.749$ & $.806$ & $.433$ & $.846$ & $.779$ & $\mathbf{.881}$ \\
\hline
\noalign{\vskip 1mm}
\multirow{5}{*}{SEDF-SC} & USleep$_{EEG}$ & $.733$ & $.981$ & $.322$ & $.821$ & $.692$ & $.847$ \\
& SOMNUS & $.748$ & $.983$ & $.336$ & $\mathbf{.835}$ & $\mathbf{.713}$ & $.871$ \\ 
& SleepFM & $.604$ & $.846$ & $.344$ & $.699$ & $.587$ & $.546$ \\
& NAPS$_P$ & $\mathbf{.772}$ & $\mathbf{.985}$ & $\mathbf{.483}$ & $.826$ & $.694$ & $.870$ \\
& NAPS$_R$ & $.752$ & $\mathbf{.985}$ & $.447$ & $.796$ & $.650$ & $\mathbf{.879}$ \\
\hline
\noalign{\vskip 1mm}
\multirow{5}{*}{SEDF-ST} & DeepResNet$_{EEG}$ & $.780$ & $.840$ & $.496$ & $.868$ & $\mathbf{.808}$ & $.886$ \\
& SOMNUS & $.767$ & $.822$ & $.444$ & $\mathbf{.878}$ & $.793$ & $\mathbf{.897}$ \\ 
& SleepFM & $.505$ & $.458$ & $.277$ & $.699$ & $.480$ & $.614$ \\
& NAPS$_P$ & $\mathbf{.813}$ & $\mathbf{.874}$ & $\mathbf{.612}$ & $.874$ & $\mathbf{.808}$ & $\mathbf{.897}$ \\
& NAPS$_R$ & $.782$ & $.845$ & $.487$ & $.873$ & $.805$ & $.892$ \\
\hline
\end{tabular}
\end{center}
\end{table*}

\clearpage

\subsection{EMG Base Models Evaluation}\label{apd:A_emg}

Given that unimodal EMG models weren't made available previously in \cite{DeiRossi2025}, but were trained for the sole purpose of this work, we report here the performances of such models on OOD datasets, allowing comparison with the performances of EEG and EOG models reported in \cite{DeiRossi2025}, and with NAPS configurations presented in the main text. 

\begin{table*}[hbt!]
\caption{Recording-wise mean(SD) performance metrics of EMG unimodal models on BSWR test split and all OOD datasets.}
\label{tab:emg_results}
\begin{center}
\scriptsize
\begin{tabular}{c|l|cccccc}
\hline
\noalign{\vskip 1mm}
\textbf{Dataset} & \textbf{Model} & \textbf{MF1} & \textbf{F1$_{W}$} & \textbf{F1$_{N1}$} & \textbf{F1$_{N2}$} & \textbf{F1$_{N3}$} & \textbf{F1$_{REM}$} \\
\hline
\noalign{\vskip 1mm}
\multirow{3}{*}{BSWR} & U-Sleep$_{EMG}$ & $\mathbf{.536(.138)}$ & $\mathbf{.693(.202)}$ & $\mathbf{.181(.122)}$ & $\mathbf{.689(.138)}$ & .474(.292) & .648(.259) \\
& DeepResNet$_{EMG}$ & .513(.133) & .684(.203) & .112(.102) & .678(.139) & .497(.292) & .609(.264) \\ 
& SleepTransformer$_{EMG}$ & .534(.138) & .691(.199) & .125(.117) & .677(.151) & $\mathbf{.536(.293)}$ & $\mathbf{.660(.254)}$ \\
\hline\hline
\noalign{\vskip 1mm}
\multirow{3}{*}{DCSM} & U-Sleep$_{EMG}$ & $\mathbf{.612(.112)}$ & $\mathbf{.913(.077)}$ & .224(.114) & $\mathbf{.707(.128)}$ & .570(.251) & .648(.225) \\
& DeepResNet$_{EMG}$ & .573(.109) & .878(.107) & .156(.099) & .628(.135) & .518(.257) & .689(.198) \\
& SleepTransformer$_{EMG}$ & .624(.098) & .910(.078) & $\mathbf{.228(.108)}$ & .671(.125) & $\mathbf{.571(.253)}$ & $\mathbf{.744(.174)}$\\
\hline
\noalign{\vskip 1mm}
\multirow{3}{*}{DOD-H} & U-Sleep$_{EMG}$ & .584(.111) & .654(.200) & .240(.096) & .761(.087) & .611(.229) & .655(.222) \\
& DeepResNet$_{EMG}$ & .590(.087) & .649(.155) & .200(.095) & .748(.081) & $\mathbf{.659(.203)}$ & .696(.219) \\ 
& SleepTransformer$_{EMG}$ & $\mathbf{.630(.101)}$ & $\mathbf{.700(.177)}$ & $\mathbf{.277(.116)}$ & $\mathbf{.764(.104)}$ & .633(.231) & $\mathbf{.779(.147)}$ \\
\hline
\noalign{\vskip 1mm}
\multirow{3}{*}{DOD-O} & U-Sleep$_{EMG}$ & $\mathbf{.577(.123)}$ & $\mathbf{.750(.137)}$ & $\mathbf{.216(.113)}$ & $\mathbf{.725(.132)}$ & $\mathbf{.464(.270)}$ & $\mathbf{.731(.240)}$ \\
& DeepResNet$_{EMG}$ & .556(.111) & .720(.153) & .192(.112) & .723(.103) & .452(.262) & .696(.226) \\ 
& SleepTransformer$_{EMG}$ & .527(.128) & .732(.150) & .190(.105) & .721(.142) & .411(.295) & .575(.298) \\
\hline
\noalign{\vskip 1mm}
\multirow{3}{*}{PHYS} & U-Sleep$_{EMG}$ & $\mathbf{.516(.108)}$ & $\mathbf{.645(.187)}$ & .205(.111) & $\mathbf{.678(.139)}$ & $\mathbf{.371(.248)}$ & $\mathbf{.683(.227)}$ \\
& DeepResNet$_{EMG}$ & .480(.111) & .621(.183) & .164(.105) & .660(.140) & .307(.256) & .645(.233) \\ 
& SleepTransformer$_{EMG}$ & .486(.113) & .629(.187) & $\mathbf{.214(.121)}$ & .667(.141) & .338(.266) & .581(.253) \\
\hline
\noalign{\vskip 1mm}
\multirow{3}{*}{SEDF-ST} & U-Sleep$_{EMG}$ & .468(.076) & .619(.158) & $\mathbf{.164(.083)}$ & .699(.103) & .173(.227) & .675(.154) \\
& DeepResNet$_{EMG}$ & .473(.062) & $\mathbf{.658(.142)}$ & .145(.069) & $\mathbf{.700(.088)}$ & .317(.232) & .539(.210) \\ 
& SleepTransformer$_{EMG}$ & $\mathbf{.481(.073)}$ & .550(.183) & .087(.099) & .657(.089) & $\mathbf{.367(.213)}$ & $\mathbf{.737(.139)}$\\
\hline
\end{tabular}
\end{center}
\end{table*}

\subsection{Structural Ablations}\label{apd:A_architectural_ablations}

To isolate the contributions of the core structural components within NAPS, we conduct two targeted architectural ablation studies. These ablations aim to validate the necessity and advantages of the intermediate tri-axial transformer processing and the attention-based fusion module. We focus on the late fusion case, as this was found to be more reliable for out-of-domain generalization.

The first ablation, denoted as \textit{NAPS-Direct}, assesses the performance impact of bypassing the tri-axial module. In this simplified architecture, the input representations from the base models are linearly projected to a shared hidden dimension ($d_{\text{model}}$) and enriched with learnable modality embeddings. However, the tri-axial transformer encoder stack, which normally enables contextualization across time, channels, and views, is completely removed. Instead, representations are directly passed into the attention-based fusion module. Due to the flexible attention mechanism of the fusion module, this ablated architecture gracefully retains the ability to process inputs with varying modality and channel counts.

In the second ablation, denoted as \textit{NAPS-Avg}, we retain the tri-axial processing component but substitute the adaptive weighting mechanism which follows it. The input representations undergo the standard projection, positional and modality embedding, and full tri-axial processing via the transformer encoder stack. However, instead of employing the attention-based fusion layer to dynamically weight the importance of the representations obtained, the model executes a na\"ive unweighted average pooling operation which is then directly fed into the final classifier head.

The empirical results for these ablations are reported in Table \ref{tab:ablation_results} and discussed below.

\begin{table*}[hbt!]
\caption{Architectural ablation results on OOD datasets. We report the per-recording mean (SD) Macro-F1 (MF1) score. \textit{NAPS} represents the full architecture. \textit{NAPS-Direct} ablates the tri-axial processing layer. \textit{NAPS-Avg} retains tri-axial processing but replaces the attention-based fusion module with average pooling. Best results per dataset are highlighted in bold.}
\label{tab:ablation_results}
\begin{center}
\scriptsize
\begin{tabular}{l|cccccc}
\hline
\noalign{\vskip 1mm}
\textbf{Model Variant} & \textbf{DCSM} & \textbf{DOD-H} & \textbf{DOD-O} & \textbf{PHYS} & \textbf{SEDF-SC} & \textbf{SEDF-ST} \\
\hline
\noalign{\vskip 1mm}
SOMNUS & $.801(.083)$ & $.829(.062)$ & $\mathbf{.790(.083)}$ & $.689(.098)$ & $.734(.083)$ & $.746(.077)$ \\
\hline
\noalign{\vskip 1mm}
NAPS$_P$ & $\textbf{.818(.081)}$ & $\mathbf{.835(.070)}$ & $.785(.085)$ & $\mathbf{.744(.095)}$ & $\mathbf{.757(.082)}$ & $\mathbf{.798(.077)}$ \\
NAPS-Direct$_P$ & $.815(.080)$ & $.820(.074)$ & $.756(.091)$ & $.737(.097)$ & $.750(.083)$ & $.790(.076)$ \\
NAPS-Avg$_P$ & $.808(.081)$ & $.821(.069)$ & $.755(.087)$ & $.742(.094)$ & $\mathbf{.757(.081)}$ & $.797(.077)$ \\
\hline
\end{tabular}
\end{center}
\end{table*}

Overall, the integration of either the tri-axial processing or the attention-based fusion module typically yields superior performance compared to soft-voting ensembling (SOMNUS). However, the combination of both components consistently achieves the most robust results across out-of-distribution datasets. Specifically, the complete removal of the tri-axial transformer stack (\textit{NAPS-Direct}) results in a more pronounced degradation in performance, underscoring the importance of contextualizing representations across time, channels, and views prior to aggregation. 
Conversely, substituting the attention-based fusion with a simple average pooling operation (\textit{NAPS-Avg}) following tri-axial processing incurs only a minor performance penalty, if any. Nevertheless, this substitution fundamentally obscures the model's predictive routing, sacrificing the transparent attribution of importance detailed in Appendix \ref{app:interpretability}, and restricts the architecture's inherent flexibility. Consequently, employing a learnable convex combination via attention fusion introduces no practical downside, while providing essential interpretability.

\subsection{Impact of Partial Modality Availability on Performance}\label{apd:A_additional_experiments}

In this section, we report the complete performance metrics for all modality subsets across all seven evaluation datasets. Table \ref{tab:ablation_bswr} through Table \ref{tab:ablation_sedf_st} compare the best unimodal baseline, the SOMNUS ensemble, NAPS$_P$, and NAPS$_R$.

These results highlight the flexibility enabled by dynamic batching, allowing NAPS to leverage arbitrary sensor combinations without retraining. Performance consistently improves with additional modalities, yet NAPS maintains a significant advantage over the SOMNUS baseline and individual constituents even in unimodal settings, notably extracting superior utility from weaker signals like EMG. Consistent with our main findings, intermediate fusion (NAPS$_R$) excels with in-domain adaptation (BSWR), whereas late fusion (NAPS$_P$) demonstrates superior stability for challenging zero-shot generalization to unseen cohorts, mitigating the risks of negative transfer under severe domain shifts.

\begin{table*}[hbt!]
\caption{Performance comparison of the best unimodal baseline, SOMNUS, NAPS$_P$, and NAPS$_R$ across different modality subsets on the BSWR dataset. For every metric, the best result per subset is bolded, and the best result overall is underlined.}
\label{tab:ablation_bswr}
\begin{center}
\scriptsize
\begin{tabular}{c|l|cccccc}
\hline
\noalign{\vskip 1mm}
\textbf{Subset} & \textbf{Model} & \textbf{MF1} & \textbf{F1$_{W}$} & \textbf{F1$_{N1}$} & \textbf{F1$_{N2}$} & \textbf{F1$_{N3}$} & \textbf{F1$_{REM}$} \\
\hline
\noalign{\vskip 1mm}
\multirow{4}{*}{EEG} & DeepResNet$_{EEG}$ & $.692(.126)$ & $.811(.157)$ & $.402(.165)$ & $.799(.148)$ & $.627(.271)$ & $.846(.195)$ \\
& SOMNUS & $.696(.129)$ & $.812(.158)$ & $.387(.172)$ & $.805(.151)$ & $.650(.286)$ & $.851(.196)$ \\
& NAPS$_P$ & $.749(.126)$ & $.840(.141)$ & $.566(.158)$ & $.812(.157)$ & $.695(.268)$ & $.856(.186)$ \\
& NAPS$_R$ & $\textbf{.779(.104)}$ & $\textbf{.852(.130)}$ & $\textbf{.603(.149)}$ & $\textbf{.832(.116)}$ & $\underline{\textbf{.759(.230)}}$ & $\textbf{.866(.169)}$ \\
\hline
\noalign{\vskip 1mm}
\multirow{4}{*}{EOG} & SleepTransformer$_{EOG}$ & $.686(.133)$ & $.794(.169)$ & $.395(.179)$ & $.783(.150)$ & $.649(.281)$ & $.837(.216)$ \\
& SOMNUS & $.696(.136)$ & $.807(.167)$ & $.429(.172)$ & $.800(.149)$ & $.623(.301)$ & $.844(.214)$ \\
& NAPS$_P$ & $.743(.125)$ & $.827(.151)$ & $.564(.152)$ & $.808(.144)$ & $.687(.276)$ & $.850(.205)$ \\
& NAPS$_R$ & $\textbf{.761(.111)}$ & $\textbf{.840(.136)}$ & $\textbf{.567(.150)}$ & $\textbf{.825(.113)}$ & $\textbf{.720(.264)}$ & $\textbf{.866(.170)}$ \\
\hline
\noalign{\vskip 1mm}
\multirow{4}{*}{EMG} & U-Sleep$_{EMG}$ & $.536(.138)$ & $.693(.202)$ & $.181(.122)$ & $.689(.138)$ & $.474(.292)$ & $.648(.259)$ \\
& SOMNUS & $.547(.136)$ & $.722(.193)$ & $.106(.106)$ & $.711(.139)$ & $.521(.305)$ & $.692(.257)$ \\
& NAPS$_P$ & $.632(.146)$ & $.728(.192)$ & $\textbf{.429(.147)}$ & $.717(.143)$ & $\textbf{.600(.291)}$ & $.698(.257)$ \\
& NAPS$_R$ & $\textbf{.641(.150)}$ & $\textbf{.741(.189)}$ & $.420(.169)$ & $\textbf{.728(.139)}$ & $\textbf{.600(.305)}$ & $\textbf{.726(.242)}$ \\
\hline
\noalign{\vskip 1mm}
\multirow{4}{*}{EEG, EOG} & DeepResNet$_{EEG}$ & $.692(.126)$ & $.811(.157)$ & $.402(.165)$ & $.799(.148)$ & $.627(.271)$ & $.846(.195)$ \\
& SOMNUS & $.704(.127)$ & $.816(.157)$ & $.405(.175)$ & $.810(.146)$ & $.653(.284)$ & $.862(.183)$ \\
& NAPS$_P$ & $.755(.123)$ & $.841(.142)$ & $.578(.154)$ & $.817(.152)$ & $.700(.270)$ & $.863(.184)$ \\
& NAPS$_R$ & $\underline{\textbf{.784(.098)}}$ & $\textbf{.853(.128)}$ & $\underline{\textbf{.609(.144)}}$ & $\underline{\textbf{.837(.109)}}$ & $\underline{\textbf{.759(.229)}}$ & $\underline{\textbf{.878(.157)}}$ \\
\hline
\noalign{\vskip 1mm}
\multirow{4}{*}{EEG, EMG} & DeepResNet$_{EEG}$ & $.692(.126)$ & $.811(.157)$ & $.402(.165)$ & $.799(.148)$ & $.627(.271)$ & $.846(.195)$ \\
& SOMNUS & $.685(.125)$ & $.808(.160)$ & $.342(.166)$ & $.802(.149)$ & $.646(.281)$ & $.854(.188)$ \\
& NAPS$_P$ & $.752(.123)$ & $.842(.141)$ & $.566(.158)$ & $.813(.156)$ & $.703(.261)$ & $.859(.178)$ \\
& NAPS$_R$ & $\textbf{.781(.100)}$ & $\underline{\textbf{.854(.128)}}$ & $\textbf{.603(.150)}$ & $\textbf{.834(.110)}$ & $\textbf{.758(.229)}$ & $\textbf{.870(.161)}$ \\
\hline
\noalign{\vskip 1mm}
\multirow{4}{*}{EOG, EMG} & SleepTransformer$_{EOG}$ & $.686(.133)$ & $.794(.169)$ & $.395(.179)$ & $.783(.150)$ & $.649(.281)$ & $.837(.216)$ \\
& SOMNUS & $.680(.134)$ & $.802(.167)$ & $.355(.172)$ & $.794(.148)$ & $.623(.290)$ & $.850(.198)$ \\
& NAPS$_P$ & $.746(.124)$ & $.829(.148)$ & $.563(.152)$ & $.809(.144)$ & $.697(.272)$ & $.852(.201)$ \\
& NAPS$_R$ & $\textbf{.762(.111)}$ & $\textbf{.841(.135)}$ & $\textbf{.567(.148)}$ & $\textbf{.826(.112)}$ & $\textbf{.717(.270)}$ & $\textbf{.871(.164)}$ \\
\hline
\noalign{\vskip 1mm}
\multirow{4}{*}{EEG, EOG, EMG} & DeepResNet$_{EEG}$ & $.692(.126)$ & $.811(.157)$ & $.402(.165)$ & $.799(.148)$ & $.627(.271)$ & $.846(.195)$ \\
& SOMNUS & $.696(.126)$ & $.812(.159)$ & $.374(.172)$ & $.807(.146)$ & $.649(.283)$ & $.862(.183)$ \\
& NAPS$_P$ & $.756(.123)$ & $.842(.142)$ & $.576(.154)$ & $.817(.152)$ & $.705(.266)$ & $.862(.185)$ \\
& NAPS$_R$ & $\underline{\textbf{.784(.099)}}$ & $\underline{\textbf{.854(.128)}}$ & $\underline{\textbf{.609(.144)}}$ & $\underline{\textbf{.837(.109)}}$ & $\textbf{.758(.230)}$ & $\underline{\textbf{.878(.155)}}$ \\
\hline
\end{tabular}
\end{center}
\end{table*}

\begin{table*}[hbt!]
\caption{Performance comparison of the best unimodal baseline, SOMNUS, NAPS$_P$, and NAPS$_R$ across different modality subsets on the DCSM dataset. For every metric, the best result per subset is bolded, and the best result overall is underlined.}
\label{tab:ablation_dcsm}
\begin{center}
\scriptsize
\begin{tabular}{c|l|cccccc}
\hline
\noalign{\vskip 1mm}
\textbf{Subset} & \textbf{Model} & \textbf{MF1} & \textbf{F1$_{W}$} & \textbf{F1$_{N1}$} & \textbf{F1$_{N2}$} & \textbf{F1$_{N3}$} & \textbf{F1$_{REM}$} \\
\hline
\noalign{\vskip 1mm}
\multirow{4}{*}{EEG} & DeepResNet$_{EEG}$ & $.797(.086)$ & $.981(.027)$ & $.507(.147)$ & $.849(.096)$ & $.779(.207)$ & $.874(.149)$ \\
& SOMNUS & $.801(.085)$ & $.984(.022)$ & $.499(.153)$ & $\textbf{.858(.097)}$ & $.782(.206)$ & $\textbf{.886(.151)}$ \\
& NAPS$_P$ & $.815(.083)$ & $.985(.021)$ & $\textbf{.562(.140)}$ & $.846(.108)$ & $.802(.195)$ & $.883(.153)$ \\
& NAPS$_R$ & $\textbf{.816(.080)}$ & $\underline{\textbf{.986(.021)}}$ & $.553(.149)$ & $.849(.097)$ & $\textbf{.815(.180)}$ & $.885(.149)$ \\
\hline
\noalign{\vskip 1mm}
\multirow{4}{*}{EOG} & SleepTransformer$_{EOG}$ & $.785(.085)$ & $.979(.029)$ & $.454(.147)$ & $.830(.106)$ & $.787(.185)$ & $.882(.147)$ \\
& SOMNUS & $.794(.084)$ & $.981(.026)$ & $.487(.148)$ & $.844(.105)$ & $.769(.200)$ & $\textbf{.892(.139)}$ \\
& NAPS$_P$ & $\textbf{.802(.082)}$ & $\textbf{.984(.022)}$ & $\textbf{.524(.140)}$ & $.827(.113)$ & $\textbf{.794(.192)}$ & $.890(.139)$ \\
& NAPS$_R$ & $.792(.082)$ & $.981(.025)$ & $.459(.155)$ & $\textbf{.851(.094)}$ & $.781(.196)$ & $.891(.142)$ \\
\hline
\noalign{\vskip 1mm}
\multirow{4}{*}{EMG} & SleepTransformer$_{EMG}$ & $.624(.098)$ & $.910(.078)$ & $.228(.107)$ & $.671(.125)$ & $.571(.252)$ & $.744(.173)$ \\
& SOMNUS & $.637(.102)$ & $.926(.075)$ & $.191(.109)$ & $\textbf{.711(.128)}$ & $.587(.261)$ & $.776(.173)$ \\
& NAPS$_P$ & $\textbf{.673(.102)}$ & $.920(.079)$ & $\textbf{.348(.122)}$ & $.696(.135)$ & $\textbf{.636(.244)}$ & $.771(.176)$ \\
& NAPS$_R$ & $.669(.104)$ & $\textbf{.930(.078)}$ & $.305(.127)$ & $.704(.133)$ & $.616(.257)$ & $\textbf{.793(.167)}$ \\
\hline
\noalign{\vskip 1mm}
\multirow{4}{*}{EEG, EOG} & DeepResNet$_{EEG}$ & $.797(.086)$ & $.981(.027)$ & $.507(.147)$ & $.849(.096)$ & $.779(.207)$ & $.874(.149)$ \\
& SOMNUS & $.803(.084)$ & $.983(.023)$ & $.505(.153)$ & $\textbf{.858(.097)}$ & $.783(.202)$ & $.891(.146)$ \\
& NAPS$_P$ & $.817(.082)$ & $\underline{\textbf{.986(.022)}}$ & $\underline{\textbf{.564(.140)}}$ & $.845(.110)$ & $.806(.191)$ & $.892(.145)$ \\
& NAPS$_R$ & $\underline{\textbf{.819(.079)}}$ & $.985(.021)$ & $.548(.149)$ & $.857(.094)$ & $\underline{\textbf{.816(.182)}}$ & $\underline{\textbf{.894(.144)}}$ \\
\hline
\noalign{\vskip 1mm}
\multirow{4}{*}{EEG, EMG} & DeepResNet$_{EEG}$ & $.797(.086)$ & $.981(.027)$ & $.507(.147)$ & $.849(.096)$ & $.779(.207)$ & $.874(.149)$ \\
& SOMNUS & $.798(.085)$ & $.983(.022)$ & $.490(.147)$ & $\underline{\textbf{.859(.096)}}$ & $.776(.211)$ & $\textbf{.888(.150)}$ \\
& NAPS$_P$ & $.815(.083)$ & $.985(.021)$ & $\textbf{.563(.141)}$ & $.846(.107)$ & $.803(.195)$ & $.884(.154)$ \\
& NAPS$_R$ & $\textbf{.818(.080)}$ & $\underline{\textbf{.986(.020)}}$ & $.557(.149)$ & $.852(.097)$ & $\textbf{.813(.183)}$ & $.886(.149)$ \\
\hline
\noalign{\vskip 1mm}
\multirow{4}{*}{EOG, EMG} & SleepTransformer$_{EOG}$ & $.785(.085)$ & $.979(.029)$ & $.454(.147)$ & $.830(.106)$ & $.787(.185)$ & $.882(.147)$ \\
& SOMNUS & $.785(.086)$ & $.981(.024)$ & $.460(.144)$ & $.846(.101)$ & $.750(.214)$ & $.892(.139)$ \\
& NAPS$_P$ & $\textbf{.804(.081)}$ & $\textbf{.984(.022)}$ & $\textbf{.529(.140)}$ & $.831(.111)$ & $\textbf{.792(.193)}$ & $.890(.139)$ \\
& NAPS$_R$ & $.790(.084)$ & $.981(.023)$ & $.463(.154)$ & $\textbf{.851(.096)}$ & $.764(.206)$ & $\textbf{.893(.139)}$ \\
\hline
\noalign{\vskip 1mm}
\multirow{4}{*}{EEG, EOG, EMG} & DeepResNet$_{EEG}$ & $.797(.086)$ & $.981(.027)$ & $.507(.147)$ & $.849(.096)$ & $.779(.207)$ & $.874(.149)$ \\
& SOMNUS & $.801(.083)$ & $.983(.023)$ & $.497(.150)$ & $.858(.096)$ & $.778(.206)$ & $.892(.145)$ \\
& NAPS$_P$ & $.818(.081)$ & $\underline{\textbf{.986(.022)}}$ & $\underline{\textbf{.564(.139)}}$ & $.846(.109)$ & $.806(.191)$ & $.892(.143)$ \\
& NAPS$_R$ & $\underline{\textbf{.819(.079)}}$ & $.985(.021)$ & $.547(.149)$ & $\underline{\textbf{.859(.094)}}$ & $\textbf{.813(.185)}$ & $\underline{\textbf{.894(.145)}}$ \\
\hline
\end{tabular}
\end{center}
\end{table*}

\begin{table*}[hbt!]
\caption{Performance comparison of the best unimodal baseline, SOMNUS, NAPS$_P$, and NAPS$_R$ across different modality subsets on the DOD-H dataset. For every metric, the best result per subset is bolded, and the best result overall is underlined.}
\label{tab:ablation_dod_h}
\begin{center}
\scriptsize
\begin{tabular}{c|l|cccccc}
\hline
\noalign{\vskip 1mm}
\textbf{Subset} & \textbf{Model} & \textbf{MF1} & \textbf{F1$_{W}$} & \textbf{F1$_{N1}$} & \textbf{F1$_{N2}$} & \textbf{F1$_{N3}$} & \textbf{F1$_{REM}$} \\
\hline
\noalign{\vskip 1mm}
\multirow{4}{*}{EEG} & U-Sleep$_{EEG}$ & $.816(.072)$ & $.878(.085)$ & $.526(.166)$ & $.907(.051)$ & $.851(.171)$ & $.916(.073)$ \\
& SOMNUS & $.822(.065)$ & $.884(.089)$ & $.531(.165)$ & $\textbf{.909(.044)}$ & $\textbf{.862(.163)}$ & $.923(.062)$ \\
& NAPS$_P$ & $\textbf{.827(.071)}$ & $.868(.102)$ & $\textbf{.608(.160)}$ & $.897(.053)$ & $.839(.166)$ & $.920(.066)$ \\
& NAPS$_R$ & $.817(.061)$ & $\textbf{.888(.078)}$ & $.532(.168)$ & $.895(.051)$ & $.843(.163)$ & $\textbf{.927(.057)}$ \\
\hline
\noalign{\vskip 1mm}
\multirow{4}{*}{EOG} & SleepTransformer$_{EOG}$ & $.796(.157)$ & $.844(.119)$ & $.567(.186)$ & $.865(.181)$ & $.803(.228)$ & $.899(.188)$ \\
& SOMNUS & $.810(.085)$ & $.838(.137)$ & $.554(.199)$ & $.894(.064)$ & $\textbf{.835(.154)}$ & $.927(.060)$ \\
& NAPS$_P$ & $\textbf{.820(.066)}$ & $.852(.093)$ & $\textbf{.598(.172)}$ & $.899(.038)$ & $.818(.159)$ & $.933(.053)$ \\
& NAPS$_R$ & $.815(.070)$ & $\textbf{.863(.112)}$ & $.543(.174)$ & $\textbf{.903(.042)}$ & $.831(.156)$ & $\textbf{.937(.052)}$ \\
\hline
\noalign{\vskip 1mm}
\multirow{4}{*}{EMG} & SleepTransformer$_{EMG}$ & $.630(.099)$ & $.700(.173)$ & $.277(.113)$ & $.764(.101)$ & $.633(.226)$ & $.779(.144)$ \\
& SOMNUS & $.643(.100)$ & $.717(.185)$ & $.220(.088)$ & $\textbf{.808(.085)}$ & $.691(.221)$ & $.786(.154)$ \\
& NAPS$_P$ & $\textbf{.673(.092)}$ & $.712(.178)$ & $\textbf{.403(.094)}$ & $.783(.087)$ & $\textbf{.693(.198)}$ & $.781(.158)$ \\
& NAPS$_R$ & $.657(.097)$ & $\textbf{.730(.176)}$ & $.274(.101)$ & $.804(.078)$ & $.677(.235)$ & $\textbf{.800(.142)}$ \\
\hline
\noalign{\vskip 1mm}
\multirow{4}{*}{EEG, EOG} & U-Sleep$_{EEG}$ & $.816(.072)$ & $.878(.085)$ & $.526(.166)$ & $.907(.051)$ & $.851(.171)$ & $.916(.073)$ \\
& SOMNUS & $.828(.064)$ & $.886(.089)$ & $.546(.162)$ & $\textbf{.912(.043)}$ & $\textbf{.866(.161)}$ & $.930(.056)$ \\
& NAPS$_P$ & $\textbf{.834(.070)}$ & $.877(.099)$ & $\textbf{.619(.157)}$ & $.901(.049)$ & $.841(.163)$ & $.933(.054)$ \\
& NAPS$_R$ & $.822(.060)$ & $\textbf{.889(.079)}$ & $.537(.168)$ & $.901(.044)$ & $.848(.161)$ & $\textbf{.934(.054)}$ \\
\hline
\noalign{\vskip 1mm}
\multirow{4}{*}{EEG, EMG} & U-Sleep$_{EEG}$ & $.816(.072)$ & $.878(.085)$ & $.526(.166)$ & $.907(.051)$ & $.851(.171)$ & $.916(.073)$ \\
& SOMNUS & $.821(.064)$ & $.884(.088)$ & $.521(.166)$ & $\textbf{.911(.043)}$ & $\textbf{.865(.161)}$ & $.926(.057)$ \\
& NAPS$_P$ & $\textbf{.827(.071)}$ & $.869(.099)$ & $\textbf{.609(.159)}$ & $.897(.053)$ & $.839(.166)$ & $.923(.066)$ \\
& NAPS$_R$ & $.817(.061)$ & $\textbf{.887(.079)}$ & $.527(.168)$ & $.897(.049)$ & $.846(.162)$ & $\textbf{.927(.057)}$ \\
\hline
\noalign{\vskip 1mm}
\multirow{4}{*}{EOG, EMG} & SleepTransformer$_{EOG}$ & $.796(.157)$ & $.844(.119)$ & $.567(.186)$ & $.865(.181)$ & $.803(.228)$ & $.899(.188)$ \\
& SOMNUS & $.812(.065)$ & $.851(.109)$ & $.521(.178)$ & $.909(.038)$ & $.841(.156)$ & $\underline{\textbf{.938(.051)}}$ \\
& NAPS$_P$ & $\textbf{.824(.063)}$ & $.854(.096)$ & $\textbf{.610(.147)}$ & $.901(.039)$ & $.820(.162)$ & $.935(.051)$ \\
& NAPS$_R$ & $.818(.062)$ & $\textbf{.868(.098)}$ & $.531(.167)$ & $\textbf{.912(.034)}$ & $\textbf{.844(.153)}$ & $.936(.049)$ \\
\hline
\noalign{\vskip 1mm}
\multirow{4}{*}{EEG, EOG, EMG} & U-Sleep$_{EEG}$ & $.816(.072)$ & $.878(.085)$ & $.526(.166)$ & $.907(.051)$ & $.851(.171)$ & $.916(.073)$ \\
& SOMNUS & $.829(.062)$ & $.887(.085)$ & $.542(.159)$ & $\underline{\textbf{.913(.042)}}$ & $\underline{\textbf{.870(.162)}}$ & $.932(.053)$ \\
& NAPS$_P$ & $\underline{\textbf{.835(.070)}}$ & $.878(.099)$ & $\underline{\textbf{.620(.158)}}$ & $.901(.049)$ & $.840(.163)$ & $\textbf{.935(.050)}$ \\
& NAPS$_R$ & $.823(.060)$ & $\underline{\textbf{.890(.079)}}$ & $.536(.165)$ & $.903(.044)$ & $.851(.162)$ & $\textbf{.935(.052)}$ \\
\hline
\end{tabular}
\end{center}
\end{table*}

\begin{table*}[hbt!]
\caption{Performance comparison of the best unimodal baseline, SOMNUS, NAPS$_P$, and NAPS$_R$ across different modality subsets on the DOD-O dataset. For every metric, the best result per subset is bolded, and the best result overall is underlined.}
\label{tab:ablation_dod_o}
\begin{center}
\scriptsize
\begin{tabular}{c|l|cccccc}
\hline
\noalign{\vskip 1mm}
\textbf{Subset} & \textbf{Model} & \textbf{MF1} & \textbf{F1$_{W}$} & \textbf{F1$_{N1}$} & \textbf{F1$_{N2}$} & \textbf{F1$_{N3}$} & \textbf{F1$_{REM}$} \\
\hline
\noalign{\vskip 1mm}
\multirow{4}{*}{EEG} & U-Sleep$_{EEG}$ & $\textbf{.776(.082)}$ & $\textbf{.906(.076)}$ & $.496(.145)$ & $\textbf{.882(.070)}$ & $.696(.264)$ & $\textbf{.904(.099)}$ \\
& SOMNUS & $.771(.093)$ & $.902(.092)$ & $.495(.162)$ & $.865(.086)$ & $\textbf{.724(.274)}$ & $.879(.095)$ \\
& NAPS$_P$ & $.760(.090)$ & $.858(.115)$ & $\textbf{.526(.131)}$ & $.846(.086)$ & $.698(.263)$ & $.878(.106)$ \\
& NAPS$_R$ & $.709(.096)$ & $.861(.120)$ & $.483(.161)$ & $.774(.113)$ & $.586(.255)$ & $.843(.120)$ \\
\hline
\noalign{\vskip 1mm}
\multirow{4}{*}{EOG} & DeepResNet$_{EOG}$ & $.745(.079)$ & $\textbf{.896(.061)}$ & $.484(.132)$ & $.866(.071)$ & $.564(.295)$ & $.912(.075)$ \\
& SOMNUS & $\textbf{.746(.086)}$ & $\textbf{.896(.061)}$ & $\textbf{.487(.141)}$ & $\textbf{.876(.063)}$ & $.554(.333)$ & $\underline{\textbf{.915(.079)}}$ \\
& NAPS$_P$ & $.744(.091)$ & $.882(.093)$ & $.468(.137)$ & $.859(.073)$ & $\textbf{.617(.313)}$ & $.896(.137)$ \\
& NAPS$_R$ & $\textbf{.746(.087)}$ & $\textbf{.896(.061)}$ & $.474(.148)$ & $.860(.071)$ & $.602(.303)$ & $.902(.121)$ \\
\hline
\noalign{\vskip 1mm}
\multirow{4}{*}{EMG} & U-Sleep$_{EMG}$ & $.577(.122)$ & $.750(.136)$ & $.216(.112)$ & $.725(.131)$ & $.464(.267)$ & $.731(.238)$ \\
& SOMNUS & $.587(.124)$ & $\textbf{.775(.134)}$ & $.201(.121)$ & $\textbf{.767(.140)}$ & $.459(.287)$ & $\textbf{.734(.250)}$ \\
& NAPS$_P$ & $\textbf{.600(.137)}$ & $.729(.172)$ & $\textbf{.304(.111)}$ & $.737(.131)$ & $\textbf{.507(.297)}$ & $.728(.248)$ \\
& NAPS$_R$ & $.559(.128)$ & $.774(.136)$ & $.151(.133)$ & $.765(.136)$ & $.395(.278)$ & $.708(.263)$ \\
\hline
\noalign{\vskip 1mm}
\multirow{4}{*}{EEG, EOG} & U-Sleep$_{EEG}$ & $.776(.082)$ & $.906(.076)$ & $.496(.145)$ & $\textbf{.882(.070)}$ & $.696(.264)$ & $.904(.099)$ \\
& SOMNUS & $\underline{\textbf{.790(.084)}}$ & $\textbf{.910(.075)}$ & $.517(.152)$ & $\textbf{.882(.076)}$ & $\underline{\textbf{.738(.266)}}$ & $.909(.077)$ \\
& NAPS$_P$ & $.779(.085)$ & $.877(.104)$ & $\underline{\textbf{.533(.133)}}$ & $.862(.080)$ & $.719(.259)$ & $\textbf{.911(.078)}$ \\
& NAPS$_R$ & $.742(.093)$ & $.884(.097)$ & $.504(.165)$ & $.807(.105)$ & $.622(.259)$ & $.900(.071)$ \\
\hline
\noalign{\vskip 1mm}
\multirow{4}{*}{EEG, EMG} & U-Sleep$_{EEG}$ & $\textbf{.776(.082)}$ & $.906(.076)$ & $.496(.145)$ & $\textbf{.882(.070)}$ & $.696(.264)$ & $\textbf{.904(.099)}$ \\
& SOMNUS & $\textbf{.776(.091)}$ & $\textbf{.908(.082)}$ & $.487(.161)$ & $.872(.081)$ & $\textbf{.729(.272)}$ & $.895(.090)$ \\
& NAPS$_P$ & $.765(.088)$ & $.862(.110)$ & $\textbf{.528(.132)}$ & $.850(.083)$ & $.701(.262)$ & $.892(.093)$ \\
& NAPS$_R$ & $.720(.094)$ & $.871(.111)$ & $.483(.163)$ & $.788(.107)$ & $.603(.258)$ & $.863(.095)$ \\
\hline
\noalign{\vskip 1mm}
\multirow{4}{*}{EOG, EMG} & DeepResNet$_{EOG}$ & $.745(.079)$ & $.896(.061)$ & $.484(.132)$ & $.866(.071)$ & $.564(.295)$ & $\textbf{.912(.075)}$ \\
& SOMNUS & $.746(.084)$ & $.895(.061)$ & $.466(.138)$ & $\textbf{.882(.053)}$ & $.576(.330)$ & $.907(.127)$ \\
& NAPS$_P$ & $\textbf{.749(.093)}$ & $.879(.099)$ & $.476(.139)$ & $.864(.069)$ & $\textbf{.636(.312)}$ & $.896(.137)$ \\
& NAPS$_R$ & $.738(.090)$ & $\textbf{.905(.056)}$ & $\textbf{.485(.145)}$ & $.871(.061)$ & $.523(.342)$ & $.904(.138)$ \\
\hline
\noalign{\vskip 1mm}
\multirow{4}{*}{EEG, EOG, EMG} & U-Sleep$_{EEG}$ & $.776(.082)$ & $.906(.076)$ & $.496(.145)$ & $.882(.070)$ & $.696(.264)$ & $.904(.099)$ \\
& SOMNUS & $\underline{\textbf{.790(.083)}}$ & $\underline{\textbf{.913(.068)}}$ & $.513(.152)$ & $\underline{\textbf{.885(.072)}}$ & $\textbf{.735(.268)}$ & $.912(.078)$ \\
& NAPS$_P$ & $.785(.085)$ & $.879(.103)$ & $\underline{\textbf{.533(.134)}}$ & $.864(.079)$ & $.721(.259)$ & $\textbf{.913(.078)}$ \\
& NAPS$_R$ & $.750(.091)$ & $.894(.087)$ & $.505(.167)$ & $.818(.100)$ & $.636(.260)$ & $.906(.069)$ \\
\hline
\end{tabular}
\end{center}
\end{table*}

\begin{table*}[hbt!]
\caption{Performance comparison of the best unimodal baseline, SOMNUS, NAPS$_P$, and NAPS$_R$ across different modality subsets on the PHYS dataset. For every metric, the best result per subset is bolded, and the best result overall is underlined.}
\label{tab:ablation_phys}
\begin{center}
\scriptsize
\begin{tabular}{c|l|cccccc}
\hline
\noalign{\vskip 1mm}
\textbf{Subset} & \textbf{Model} & \textbf{MF1} & \textbf{F1$_{W}$} & \textbf{F1$_{N1}$} & \textbf{F1$_{N2}$} & \textbf{F1$_{N3}$} & \textbf{F1$_{REM}$} \\
\hline
\noalign{\vskip 1mm}
\multirow{4}{*}{EEG} & DeepResNet$_{EEG}$ & $.687(.097)$ & $.744(.159)$ & $.358(.153)$ & $.832(.106)$ & $.682(.247)$ & $.837(.173)$ \\
& SOMNUS & $.690(.100)$ & $.742(.161)$ & $.337(.156)$ & $\textbf{.834(.110)}$ & $.711(.242)$ & $\textbf{.843(.172)}$ \\
& NAPS$_P$ & $\textbf{.740(.098)}$ & $\textbf{.791(.150)}$ & $\textbf{.531(.140)}$ & $.826(.111)$ & $\underline{\textbf{.722(.239)}}$ & $.842(.170)$ \\
& NAPS$_R$ & $.711(.098)$ & $.761(.157)$ & $.424(.155)$ & $.825(.109)$ & $.714(.238)$ & $\textbf{.843(.169)}$ \\
\hline
\noalign{\vskip 1mm}
\multirow{4}{*}{EOG} & SleepTransformer$_{EOG}$ & $.663(.102)$ & $.733(.165)$ & $.408(.148)$ & $.790(.111)$ & $.563(.276)$ & $.828(.179)$ \\
& SOMNUS & $.673(.103)$ & $.740(.164)$ & $.412(.152)$ & $\textbf{.815(.108)}$ & $.562(.293)$ & $.846(.172)$ \\
& NAPS$_P$ & $\textbf{.713(.100)}$ & $\textbf{.786(.152)}$ & $\textbf{.518(.138)}$ & $.799(.116)$ & $\textbf{.626(.278)}$ & $.841(.172)$ \\
& NAPS$_R$ & $.691(.099)$ & $.757(.157)$ & $.434(.149)$ & $\textbf{.815(.105)}$ & $.611(.278)$ & $\textbf{.848(.166)}$ \\
\hline
\noalign{\vskip 1mm}
\multirow{4}{*}{EMG} & U-Sleep$_{EMG}$ & $.516(.108)$ & $.645(.187)$ & $.205(.111)$ & $.678(.139)$ & $.371(.248)$ & $.683(.227)$ \\
& SOMNUS & $.517(.109)$ & $.666(.183)$ & $.169(.112)$ & $\textbf{.710(.137)}$ & $.336(.271)$ & $\textbf{.706(.224)}$ \\
& NAPS$_P$ & $\textbf{.581(.118)}$ & $\textbf{.674(.180)}$ & $\textbf{.419(.137)}$ & $.693(.142)$ & $\textbf{.424(.279)}$ & $.696(.230)$ \\
& NAPS$_R$ & $.491(.115)$ & $.667(.182)$ & $.156(.161)$ & $.709(.134)$ & $.241(.265)$ & $.675(.247)$ \\
\hline
\noalign{\vskip 1mm}
\multirow{4}{*}{EEG, EOG} & DeepResNet$_{EEG}$ & $.687(.097)$ & $.744(.159)$ & $.358(.153)$ & $.832(.106)$ & $.682(.247)$ & $.837(.173)$ \\
& SOMNUS & $.693(.099)$ & $.743(.161)$ & $.349(.157)$ & $\underline{\textbf{.837(.107)}}$ & $.704(.248)$ & $.847(.170)$ \\
& NAPS$_P$ & $\textbf{.743(.096)}$ & $\textbf{.792(.149)}$ & $\textbf{.537(.139)}$ & $.829(.109)$ & $\textbf{.721(.240)}$ & $.848(.167)$ \\
& NAPS$_R$ & $.715(.097)$ & $.762(.156)$ & $.432(.156)$ & $.829(.106)$ & $.713(.241)$ & $\textbf{.850(.165)}$ \\
\hline
\noalign{\vskip 1mm}
\multirow{4}{*}{EEG, EMG} & DeepResNet$_{EEG}$ & $.687(.097)$ & $.744(.159)$ & $.358(.153)$ & $.832(.106)$ & $.682(.247)$ & $.837(.173)$ \\
& SOMNUS & $.687(.099)$ & $.742(.161)$ & $.325(.157)$ & $\textbf{.835(.109)}$ & $.705(.246)$ & $.844(.171)$ \\
& NAPS$_P$ & $\textbf{.742(.097)}$ & $\textbf{.792(.149)}$ & $\textbf{.534(.139)}$ & $.827(.111)$ & $\underline{\textbf{.722(.239)}}$ & $.844(.168)$ \\
& NAPS$_R$ & $.707(.098)$ & $.759(.158)$ & $.406(.158)$ & $.830(.107)$ & $.707(.243)$ & $\textbf{.845(.168)}$ \\
\hline
\noalign{\vskip 1mm}
\multirow{4}{*}{EOG, EMG} & SleepTransformer$_{EOG}$ & $.663(.102)$ & $.733(.165)$ & $.408(.148)$ & $.790(.111)$ & $.563(.276)$ & $.828(.179)$ \\
& SOMNUS & $.643(.099)$ & $.730(.167)$ & $.337(.153)$ & $.810(.110)$ & $.494(.296)$ & $.849(.164)$ \\
& NAPS$_P$ & $\textbf{.718(.097)}$ & $\underline{\textbf{.793(.147)}}$ & $\textbf{.535(.136)}$ & $.805(.113)$ & $\textbf{.621(.276)}$ & $.844(.169)$ \\
& NAPS$_R$ & $.664(.101)$ & $.753(.157)$ & $.394(.159)$ & $\textbf{.817(.105)}$ & $.507(.305)$ & $\underline{\textbf{.853(.162)}}$ \\
\hline
\noalign{\vskip 1mm}
\multirow{4}{*}{EEG, EOG, EMG} & DeepResNet$_{EEG}$ & $.687(.097)$ & $.744(.159)$ & $.358(.153)$ & $.832(.106)$ & $.682(.247)$ & $.837(.173)$ \\
& SOMNUS & $.689(.098)$ & $.742(.161)$ & $.338(.158)$ & $\underline{\textbf{.837(.107)}}$ & $.697(.251)$ & $.848(.168)$ \\
& NAPS$_P$ & $\underline{\textbf{.744(.095)}}$ & $\underline{\textbf{.793(.148)}}$ & $\underline{\textbf{.538(.138)}}$ & $.830(.108)$ & $\textbf{.721(.240)}$ & $.848(.166)$ \\
& NAPS$_R$ & $.711(.096)$ & $.760(.157)$ & $.417(.157)$ & $.833(.105)$ & $.705(.246)$ & $\textbf{.851(.164)}$ \\
\hline
\end{tabular}
\end{center}
\end{table*}

\begin{table*}[hbt!]
\caption{Performance comparison of the best unimodal baseline, SOMNUS, NAPS$_P$, and NAPS$_R$ across different modality subsets on the SEDF-SC dataset. For every metric, the best result per subset is bolded, and the best result overall is underlined.}
\label{tab:ablation_sedf_sc}
\begin{center}
\scriptsize
\begin{tabular}{c|l|cccccc}
\hline
\noalign{\vskip 1mm}
\textbf{Subset} & \textbf{Model} & \textbf{MF1} & \textbf{F1$_{W}$} & \textbf{F1$_{N1}$} & \textbf{F1$_{N2}$} & \textbf{F1$_{N3}$} & \textbf{F1$_{REM}$} \\
\hline
\noalign{\vskip 1mm}
\multirow{4}{*}{EEG} & U-Sleep$_{EEG}$ & $.720(.090)$ & $.981(.014)$ & $.342(.130)$ & $.814(.097)$ & $.602(.287)$ & $.845(.114)$ \\
& SOMNUS & $.726(.086)$ & $.981(.019)$ & $.343(.139)$ & $\textbf{.822(.089)}$ & $\textbf{.606(.283)}$ & $\textbf{.862(.100)}$ \\
& NAPS$_P$ & $\textbf{.750(.084)}$ & $\textbf{.984(.017)}$ & $\textbf{.470(.124)}$ & $.814(.093)$ & $.600(.288)$ & $.861(.101)$ \\
& NAPS$_R$ & $.730(.086)$ & $\textbf{.984(.014)}$ & $.452(.122)$ & $.773(.108)$ & $.551(.286)$ & $\textbf{.862(.098)}$ \\
\hline
\noalign{\vskip 1mm}
\multirow{4}{*}{EOG} & DeepResNet$_{EOG}$ & $.676(.094)$ & $.935(.063)$ & $.410(.132)$ & $.797(.093)$ & $.534(.291)$ & $.680(.172)$ \\
& SOMNUS & $.712(.093)$ & $.974(.027)$ & $.379(.135)$ & $\textbf{.811(.100)}$ & $.547(.283)$ & $.827(.133)$ \\
& NAPS$_P$ & $\textbf{.734(.094)}$ & $.976(.024)$ & $\textbf{.468(.136)}$ & $.796(.112)$ & $.545(.288)$ & $\textbf{.854(.121)}$ \\
& NAPS$_R$ & $.718(.093)$ & $\textbf{.978(.024)}$ & $.403(.120)$ & $.790(.109)$ & $\textbf{.554(.290)}$ & $.843(.125)$ \\
\hline
\noalign{\vskip 1mm}
\multirow{4}{*}{EEG, EOG} & U-Sleep$_{EEG}$ & $.720(.090)$ & $.981(.014)$ & $.342(.130)$ & $.814(.097)$ & $.602(.287)$ & $.845(.114)$ \\
& SOMNUS & $.734(.083)$ & $.982(.018)$ & $.358(.138)$ & $\underline{\textbf{.832(.083)}}$ & $\underline{\textbf{.611(.279)}}$ & $.870(.094)$ \\
& NAPS$_P$ & $\underline{\textbf{.757(.082)}}$ & $\underline{\textbf{.985(.016)}}$ & $\underline{\textbf{.487(.124)}}$ & $.822(.087)$ & $.597(.291)$ & $.871(.095)$ \\
& NAPS$_R$ & $.739(.083)$ & $\underline{\textbf{.985(.011)}}$ & $.454(.120)$ & $.791(.098)$ & $.563(.290)$ & $\underline{\textbf{.876(.084)}}$ \\
\hline
\end{tabular}
\end{center}
\end{table*}

\begin{table*}[t!]
\caption{Performance comparison of the best unimodal baseline, SOMNUS, NAPS$_P$, and NAPS$_R$ across different modality subsets on the SEDF-ST dataset. For every metric, the best result per subset is bolded, and the best result overall is underlined.}
\label{tab:ablation_sedf_st}
\begin{center}
\scriptsize
\begin{tabular}{c|l|cccccc}
\hline
\noalign{\vskip 1mm}
\textbf{Subset} & \textbf{Model} & \textbf{MF1} & \textbf{F1$_{W}$} & \textbf{F1$_{N1}$} & \textbf{F1$_{N2}$} & \textbf{F1$_{N3}$} & \textbf{F1$_{REM}$} \\
\hline
\noalign{\vskip 1mm}
\multirow{4}{*}{EEG} & DeepResNet$_{EEG}$ & $.764(.074)$ & $.814(.105)$ & $.508(.158)$ & $.863(.062)$ & $.746(.232)$ & $.891(.085)$ \\
& SOMNUS & $.767(.075)$ & $.813(.107)$ & $.508(.159)$ & $\textbf{.872(.060)}$ & $.744(.230)$ & $\textbf{.897(.084)}$ \\
& NAPS$_P$ & $\textbf{.795(.077)}$ & $\textbf{.851(.095)}$ & $\textbf{.606(.154)}$ & $.867(.061)$ & $\textbf{.750(.232)}$ & $\textbf{.897(.085)}$ \\
& NAPS$_R$ & $.781(.075)$ & $.843(.098)$ & $.564(.158)$ & $.856(.061)$ & $.748(.230)$ & $.891(.088)$ \\
\hline
\noalign{\vskip 1mm}
\multirow{4}{*}{EOG} & U-Sleep$_{EOG}$ & $.712(.067)$ & $.748(.130)$ & $.410(.121)$ & $.846(.057)$ & $\textbf{.688(.237)}$ & $.869(.077)$ \\
& SOMNUS & $.722(.072)$ & $.758(.128)$ & $.436(.148)$ & $\textbf{.855(.054)}$ & $.664(.263)$ & $\textbf{.894(.071)}$ \\
& NAPS$_P$ & $\textbf{.762(.068)}$ & $\textbf{.825(.089)}$ & $\textbf{.558(.127)}$ & $.844(.057)$ & $\textbf{.688(.250)}$ & $\textbf{.894(.076)}$ \\
& NAPS$_R$ & $.702(.062)$ & $.752(.118)$ & $.341(.139)$ & $\textbf{.855(.059)}$ & $.666(.242)$ & $.893(.076)$ \\
\hline
\noalign{\vskip 1mm}
\multirow{4}{*}{EMG} & SleepTransformer$_{EMG}$ & $.481(.072)$ & $.550(.181)$ & $.087(.097)$ & $.657(.088)$ & $.367(.210)$ & $\textbf{.737(.137)}$ \\
& SOMNUS & $.500(.069)$ & $.651(.158)$ & $.102(.082)$ & $\textbf{.737(.087)}$ & $.272(.263)$ & $.727(.160)$ \\
& NAPS$_P$ & $\textbf{.600(.081)}$ & $\textbf{.717(.125)}$ & $\textbf{.438(.116)}$ & $.711(.082)$ & $\textbf{.397(.248)}$ & $.730(.156)$ \\
& NAPS$_R$ & $.515(.076)$ & $.674(.142)$ & $.108(.081)$ & $.736(.086)$ & $.351(.257)$ & $.696(.183)$ \\
\hline
\noalign{\vskip 1mm}
\multirow{4}{*}{EEG, EOG} & DeepResNet$_{EEG}$ & $.764(.074)$ & $.814(.105)$ & $.508(.158)$ & $.863(.062)$ & $\textbf{.746(.232)}$ & $.891(.085)$ \\
& SOMNUS & $.761(.074)$ & $.803(.105)$ & $.497(.148)$ & $\underline{\textbf{.873(.057)}}$ & $.731(.237)$ & $\underline{\textbf{.902(.080)}}$ \\
& NAPS$_P$ & $\textbf{.797(.078)}$ & $\underline{\textbf{.853(.095)}}$ & $\textbf{.616(.152)}$ & $.869(.058)$ & $.745(.235)$ & $.900(.081)$ \\
& NAPS$_R$ & $.769(.074)$ & $.824(.104)$ & $.517(.153)$ & $.865(.055)$ & $.744(.232)$ & $.895(.084)$ \\
\hline
\noalign{\vskip 1mm}
\multirow{4}{*}{EEG, EMG} & DeepResNet$_{EEG}$ & $.764(.074)$ & $.814(.105)$ & $.508(.158)$ & $.863(.062)$ & $.746(.232)$ & $.891(.085)$ \\
& SOMNUS & $.748(.070)$ & $.794(.110)$ & $.453(.149)$ & $\textbf{.871(.060)}$ & $.725(.225)$ & $.896(.083)$ \\
& NAPS$_P$ & $\textbf{.795(.077)}$ & $\textbf{.852(.096)}$ & $\textbf{.607(.154)}$ & $.868(.061)$ & $.748(.232)$ & $\textbf{.898(.084)}$ \\
& NAPS$_R$ & $.779(.075)$ & $.838(.102)$ & $.551(.155)$ & $.864(.058)$ & $\underline{\textbf{.751(.223)}}$ & $.892(.087)$ \\
\hline
\noalign{\vskip 1mm}
\multirow{4}{*}{EOG, EMG} & U-Sleep$_{EOG}$ & $.712(.067)$ & $.748(.130)$ & $.410(.121)$ & $.846(.057)$ & $\textbf{.688(.237)}$ & $.869(.077)$ \\
& SOMNUS & $.671(.057)$ & $.725(.131)$ & $.277(.129)$ & $.847(.065)$ & $.611(.263)$ & $.892(.075)$ \\
& NAPS$_P$ & $\textbf{.764(.068)}$ & $\textbf{.822(.090)}$ & $\textbf{.565(.127)}$ & $.847(.056)$ & $.685(.250)$ & $\textbf{.896(.075)}$ \\
& NAPS$_R$ & $.677(.065)$ & $.745(.118)$ & $.288(.133)$ & $\textbf{.852(.065)}$ & $.605(.289)$ & $.895(.076)$ \\
\hline
\noalign{\vskip 1mm}
\multirow{4}{*}{EEG, EOG, EMG} & DeepResNet$_{EEG}$ & $.764(.074)$ & $.814(.105)$ & $.508(.158)$ & $.863(.062)$ & $\textbf{.746(.232)}$ & $.891(.085)$ \\
& SOMNUS & $.746(.070)$ & $.786(.110)$ & $.452(.143)$ & $\textbf{.872(.058)}$ & $.716(.233)$ & $\underline{\textbf{.902(.080)}}$ \\
& NAPS$_P$ & $\underline{\textbf{.798(.077)}}$ & $\underline{\textbf{.853(.094)}}$ & $\underline{\textbf{.618(.153)}}$ & $.870(.058)$ & $\textbf{.746(.232)}$ & $\underline{\textbf{.902(.082)}}$ \\
& NAPS$_R$ & $.766(.073)$ & $.818(.103)$ & $.503(.150)$ & $.868(.056)$ & $.742(.232)$ & $.896(.084)$ \\
\hline
\end{tabular}
\end{center}
\end{table*}

\clearpage

\subsection{Data Efficiency and Meta-Training Volume Ablation}\label{apd:A_data_ablation}

To isolate the architectural contribution of the NAPS module from the effect of data volume, we evaluate whether the out-of-domain gains are merely a byproduct of exposing the fusion module to the large BSWR dataset. Table \ref{tab:data_ablation} compares the out-of-domain performance of the standard U-Sleep$_{EEG}$ encoder trained exclusively on the NSRR datasets, the same U-Sleep$_{EEG}$ encoder retrained on the combined NSRR and BSWR corpora, the SOMNUS soft-voting ensemble, and two variants of NAPS$_P$ meta-trained on 1\% and 100\% of the BSWR data.

The results demonstrate that simply adding the $\approx 80,000$ hours of BSWR data to the pre-training corpus of a unimodal baseline (U-Sleep$_{EEG}$) does not yield meaningful improvements; on the contrary, it degrades zero-shot generalization across all evaluated external cohorts. We attribute this to negative transfer: because the BSWR dataset consists almost entirely of patients with severe sleep-wake disorders, naively adding such a massive, highly pathological dataset with the NSRR corpora can skew the model's representations rather than improve its generalizability. In contrast, training NAPS$_P$ on just 1\% of the BSWR data ($\approx 800$ hours, representing only $\approx 0.36\%$ of the base models' original pre-training corpus) to learn a principled prediction-aggregation strategy is already sufficient to outperform the SOMNUS ensemble on DCSM, PHYS, SEDF-SC, and SEDF-ST. Utilizing the full 100\% of the BSWR dataset for NAPS$_P$ further improves these gains.

\begin{table}[ht!]
\caption{Data efficiency ablation. Per-recording mean (std) Macro-F1 scores comparing the baseline U-Sleep$_{EEG}$ model trained exclusively on NSRR, U-Sleep$_{EEG}$ (w/ BSWR) retrained with both the NSRR corpora and the full BSWR dataset included in its pre-training corpus, the soft-voting ensemble (SOMNUS), and NAPS$_P$ exploiting either 1\% or 100\% of the BSWR data.}
\label{tab:data_ablation}
\begin{center}
\scriptsize
\begin{tabular}{l|ccccc}
\hline
\noalign{\vskip 1mm}
\textbf{Dataset} & \textbf{U-Sleep$_{EEG}$ (NSRR)} & \textbf{U-Sleep$_{EEG}$ (w/ BSWR)} & \textbf{SOMNUS} & \textbf{NAPS$_P$ (1\%)} & \textbf{NAPS$_P$ (100\%)} \\
\hline
\noalign{\vskip 1mm}
DCSM    & $.783(.086)$ & $.757(.078)$ & $.801(.083)$ & $.809(.079)$ & $\mathbf{.818(.081)}$ \\
DOD-H   & $.816(.072)$ & $.755(.060)$ & $.829(.062)$ & $.801(.072)$ & $\mathbf{.835(.070)}$ \\
DOD-O   & $.776(.082)$ & $.721(.080)$ & $\mathbf{.790(.083)}$ & $.728(.090)$ & $.785(.085)$ \\
PHYS    & $.687(.101)$ & $.665(.096)$ & $.689(.098)$ & $.724(.100)$ & $\mathbf{.744(.095)}$ \\
SEDF-SC & $.720(.090)$ & $.660(.087)$ & $.734(.083)$ & $.738(.086)$ & $\mathbf{.757(.082)}$ \\
SEDF-ST & $.758(.074)$ & $.720(.063)$ & $.746(.070)$ & $.789(.078)$ & $\mathbf{.798(.077)}$ \\
\hline
\end{tabular}
\end{center}
\end{table}


\subsection{Interpretability and Attention Routing}\label{app:interpretability}

Unlike simple pooling or averaging schemes, which obscure the relative contribution of each input, the attention-based Modality Fusion Layer in NAPS inherently provides a transparent window into the model's decision-making process on modality (and potentially per-channel) importance. By analyzing the learned attention weights, we show that NAPS actively grounds its predictions in established sleep physiology.

\paragraph{Aggregation of Attention Weights}
As defined in Section 3.1, the fusion layer outputs a set of attention weights $\alpha_{t,n} \in [0, 1]$ that form a valid probability distribution over the $N$ representations at each time step $t$. To quantify the importance of different modalities, we define $\mathcal{R}_{m_k}$ as the set of representation indices $n$ belonging to a specific modality $m_k$. The total number of representations for this modality is $|\mathcal{R}_{m_k}| = C_{m_k} \cdot B_{m_k}$. We aggregate the attention weights using:

\begin{enumerate}
    \item \textit{Total Attention:} The cumulative weight for a given modality, representing its overall influence on the final prediction:
    $$S_{t,m_k} = \sum_{n \in \mathcal{R}_{m_k}} \alpha_{t,n}$$
    \item \textit{Per-Channel Mean Attention:} We isolate the intrinsic importance of the modality per channel by computing the mean weight per individual representation:
    $$M_{t,m_k} = \frac{1}{|\mathcal{R}_{m_k}|} \sum_{n \in \mathcal{R}_{m_k}} \alpha_{t,n} = \frac{S_{t,{m_k}}}{|\mathcal{R}_{m_k}|}$$
\end{enumerate}

\paragraph{Global Physiological Alignment}

To evaluate global modality importance across multiple recordings, we analyzed the attention weights across the entire BSWR test split (Tables \ref{tab:global_sum} and \ref{tab:global_mean}). 
The results reveal an intuitive alignment between the learned routing strategies of NAPS and the human scoring criteria:

\begin{itemize}
    \item \textit{Deep Sleep Progression:} During N2 and N3 sleep, attention is firmly dominated by the EEG modality, reflecting the physiological reliance on sleep spindles, K-complexes, and slow-wave activity. The total EEG attention steadily increases with sleep depth, peaking at an average sum weight of 0.639 in N3.
    \item \textit{The REM Reversal:} During Rapid Eye Movement (REM) sleep, NAPS learns to dynamically shift its primary focus to the ocular sensors. The EOG per-channel mean attention surges to 0.048, completely overtaking the EEG mean (0.028). Consequently, the total EOG contribution (0.485) overtakes the total EEG contribution (0.449) in REM stages. 
    \item \textit{Down-weighting of EMG:} EMG weights remain consistently lower than EEG and EOG across all stages, when all modalities are considered. Crucially, this is a desirable learned behavior. When fusing highly capable base encoders that can confidently predict most stages using brain and eye activity alone, forcing equal contribution (as in simple averaging ensembles) has the potential to dilute overall performance by over-relying on weaker signals. Instead, NAPS learns to appropriately down-weight EMG, utilizing it as a secondary verification signal rather than a primary driver.
\end{itemize}

\begin{table}[h]
\centering
\caption{Modality Attention Weights (Sum) when using all modalities. Mean (STD) across 424 BSWR test recordings.}
\label{tab:global_sum}
\begin{tabular}{l c c c}
\toprule
\textbf{Stage} & \textbf{EEG} & \textbf{EOG} & \textbf{EMG} \\
\midrule
Wake & \textbf{0.563 (0.044)} & 0.358 (0.043) & 0.080 (0.021) \\
N1   & \textbf{0.572 (0.040)} & 0.364 (0.041) & 0.065 (0.016) \\
N2   & \textbf{0.607 (0.039)} & 0.322 (0.039) & 0.072 (0.019) \\
N3   & \textbf{0.639 (0.043)} & 0.273 (0.042) & 0.088 (0.025) \\
REM  & 0.449 (0.057) & \textbf{0.485 (0.056)} & 0.066 (0.019) \\
\bottomrule
\end{tabular}
\end{table}

\begin{table}[h]
\centering
\caption{Per-Channel Attention Weights (Mean) when using all modalities. Mean (STD) across 424 BSWR test recordings.}
\label{tab:global_mean}
\begin{tabular}{l c c c}
\toprule
\textbf{Stage} & \textbf{EEG} & \textbf{EOG} & \textbf{EMG} \\
\midrule
Wake & \textbf{0.035 (0.009)} & \textbf{0.035 (0.009)} & 0.014 (0.005) \\
N1   & \textbf{0.036 (0.009)} & \textbf{0.036 (0.010)} & 0.011 (0.004) \\
N2   & \textbf{0.038 (0.010)} & 0.032 (0.008) & 0.013 (0.004) \\
N3   & \textbf{0.040 (0.010)} & 0.027 (0.007) & 0.015 (0.005) \\
REM  & 0.028 (0.009) & \textbf{0.048 (0.013)} & 0.011 (0.004) \\
\bottomrule
\end{tabular}
\end{table}

\paragraph{Attention Weighting Under Corruption}

\begin{figure}[h]
    \centering
    \includegraphics[width=\textwidth]{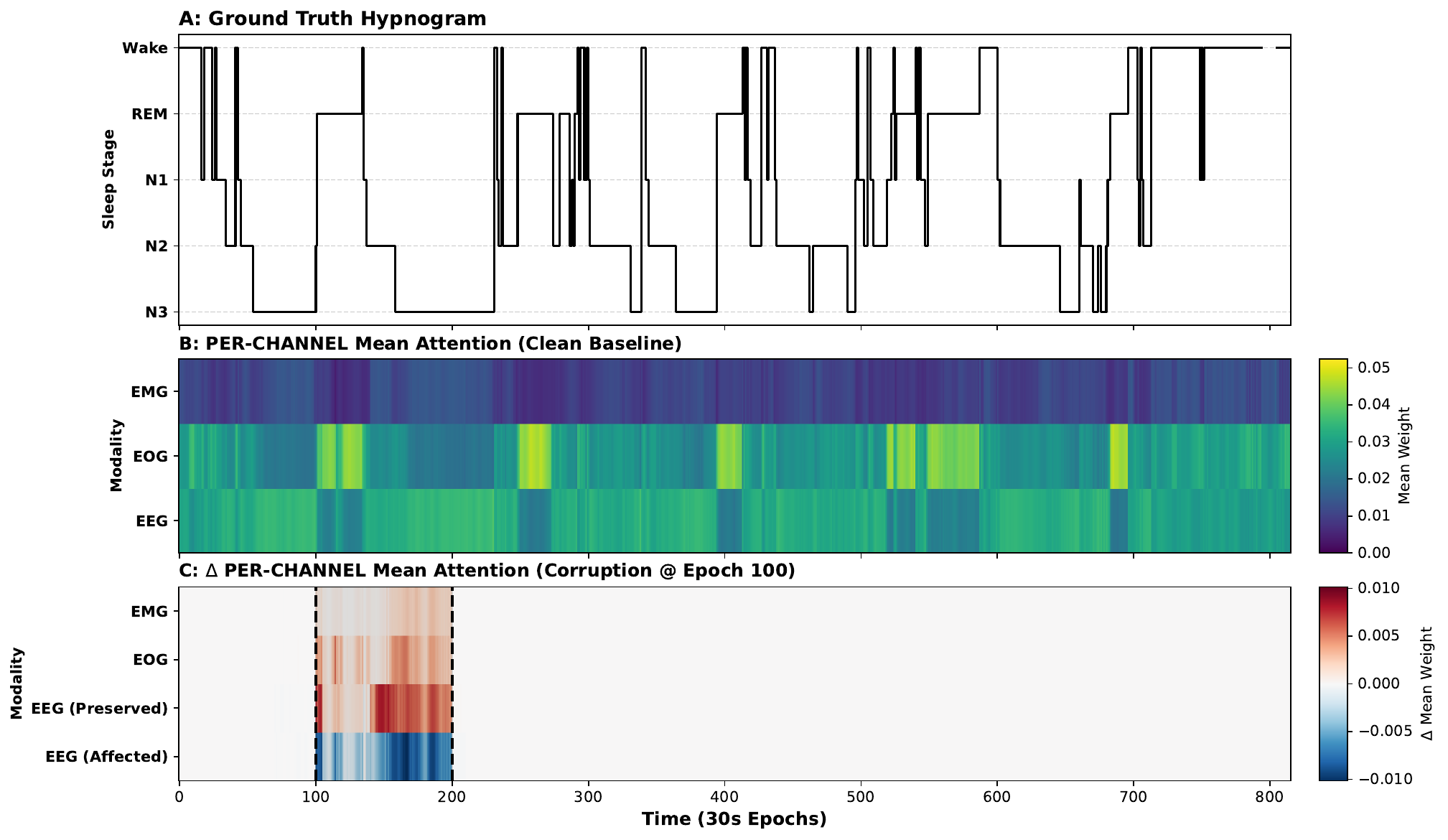}
    \caption{Scenario 1: Artificial corruption of 4 EEG channels for 100 epochs. The delta heatmap (subplot C) shows NAPS successfully routing attention away from the affected EEG channels (blue) and heavily up-weighting the preserved EEG channels (red) to compensate.}
    \label{fig:eeg_drop}
\end{figure}

To qualitatively evaluate the robustness of the Modality Fusion Layer, we artificially induce sensor failures during inference. For a continuous window of 100 epochs, we override target channels with uninformative distributions derived from the base predictors. We then analyze the resulting shift in mean attention ($\Delta = \text{after corruption} - \text{clean}$) across three distinct scenarios:

\textit{Scenario 1: Partial Sensor Failure.} When a subset of EEG electrodes are artificially corrupted (Figure \ref{fig:eeg_drop}), the attention weights for the affected channels drop significantly. Rather than indiscriminately transferring this lost attention entirely to other modalities, NAPS exhibits robust intra-modality routing; it compensates primarily by heavily up-weighting the remaining preserved EEG channels, alongside a partial increase in the other modalities. 

\begin{figure}[h]
    \centering
    \includegraphics[width=\textwidth]{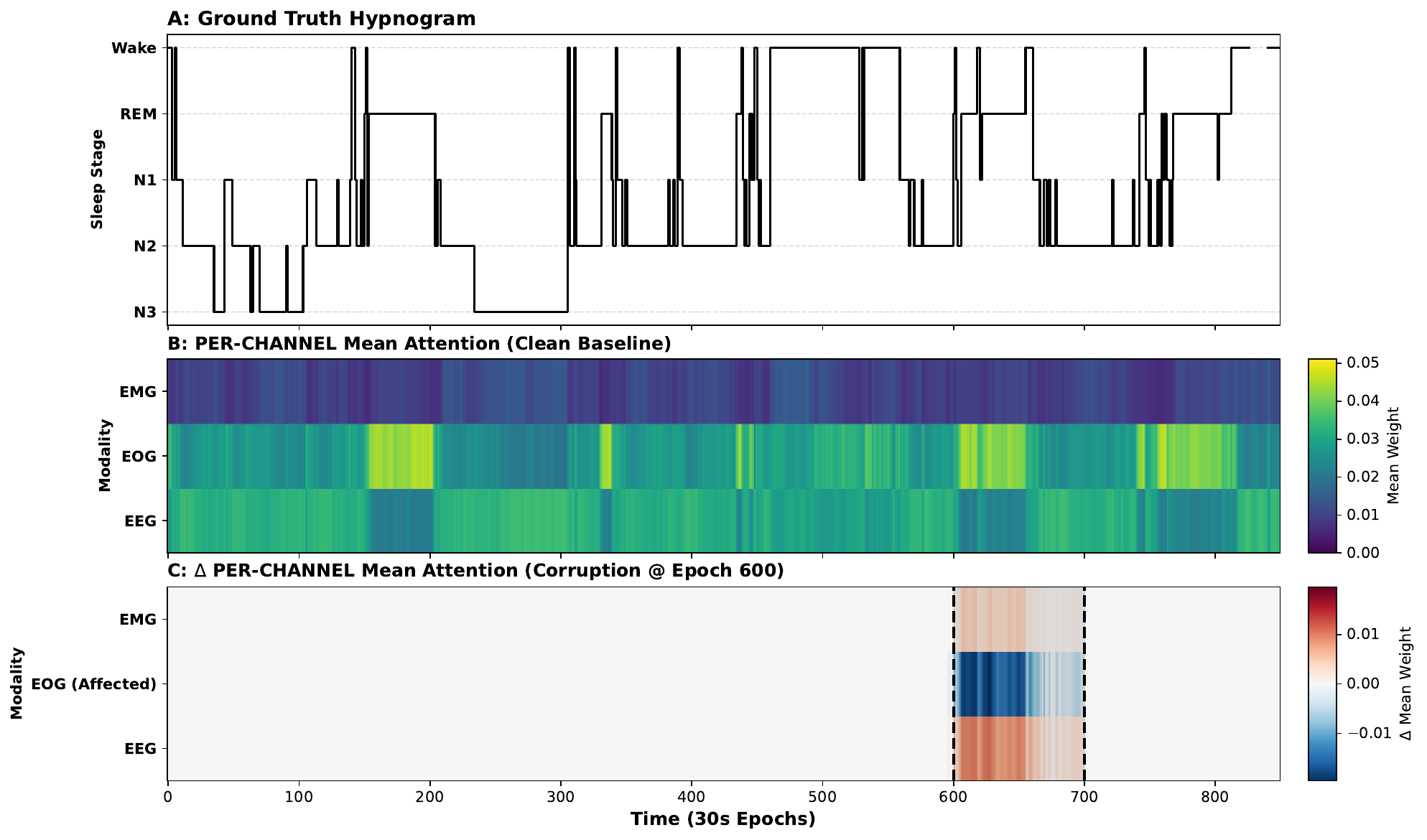}
    \caption{Scenario 2: Total failure of the EOG modality for 100 epochs. Denied access to ocular data, the network shifts attention onto both EEG and EMG, compensating for the compromised modality.}
    \label{fig:eog_drop}
\end{figure}

\textit{Scenario 2: Total Sensor Failure.} When the entire EOG modality is corrupted (Figure \ref{fig:eog_drop}), NAPS instantly recognizes the absence of reliable ocular data. It redistributes the attention mass to the surviving EEG and EMG modalities, proportionally favoring the stronger predictive signal.

\begin{figure}[h]
    \centering
    \includegraphics[width=\textwidth]{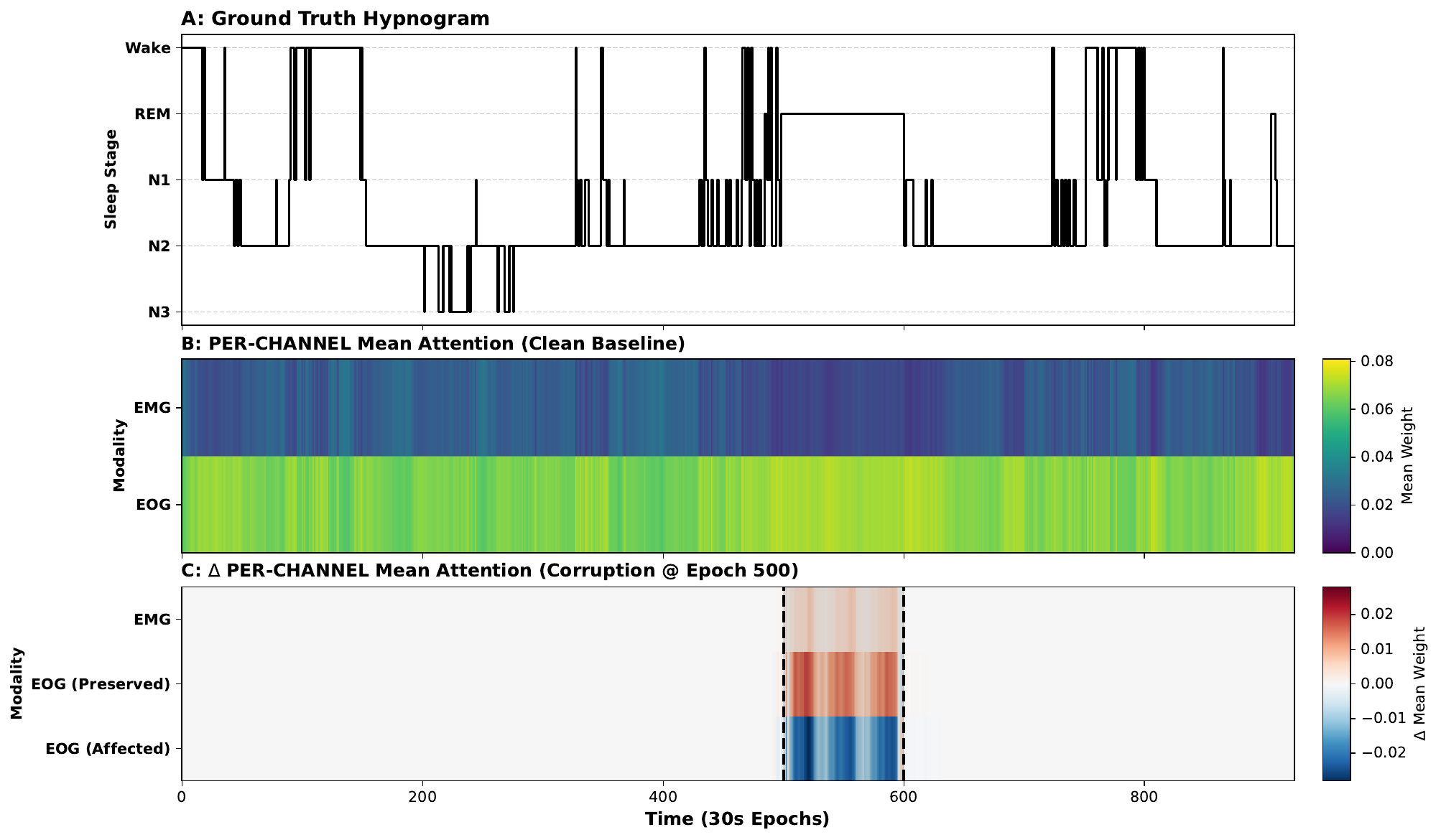}
    \caption{Scenario 3: Inference performed entirely without EEG. EOG correctly acts as the primary modality (subplot B). When 2 EOG channels are corrupted (subplot C), the network performs intra-modality routing among the remaining EOG channels while relying on EMG as a secondary signal.}
    \label{fig:no_eeg_drop}
\end{figure}

\textit{Scenario 3: Modality Absence and Intra-Modality Routing.} Lastly, we consider a recording where the EEG modality is excluded entirely (Figure \ref{fig:no_eeg_drop}). In this regime, EOG correctly assumes the role of the primary modality. When 2 EOG channels are subsequently corrupted, the network dynamically routes attention away from the affected channels and onto the preserved EOG sensors, while consistently maintaining EMG as a stable secondary backup.

\end{document}